\documentclass{IEEEtran}
\newcommand{\eat}[1]{}
\usepackage{epsfig}
\usepackage{subfigure}
\usepackage{latexsym}
\usepackage{amssymb}
\usepackage{amsmath}
\usepackage{url}
\usepackage{colortbl}

\usepackage{float}
\usepackage{graphicx}
\usepackage[boxruled]{algorithm2e}
\usepackage{algorithmic}
\usepackage{amsmath}
\usepackage{pdfsync}
\usepackage{booktabs}
\usepackage{multirow}
\usepackage{textcomp}
\newtheorem{definition}{Definition}

\newcommand{\yu}{\textcolor{blue}{YU:}\textcolor{blue}}
\begin{document}
	
	\title{RUM: network Representation learning throUgh Multi-level structural information preservation}

		\author{
			Yanlei Yu$^*$$^\dag$, Zhiwu Lu$^*$$^\dag$, $\diamond$Jiajun Liu$^*$$^\dag$, Guoping Zhao$^*$$^\dag$, Ji-Rong Wen$^*$$^\dag$, Kai Zheng$^\diamond$\\
			{$^*$Renmin University of China, \\\{yuyanlei1994,luzhiwu,jiajunliu,guopingzhao,jrwen\}@ruc.edu.cn}\\
			{$^\diamond$University of Electronic
Science and Technology of China},\\
			{zhengkai@uestc.edu.cn}\\
			{$^\dag$Beijing Key Laboratory of Big Data Management and Analysis Methods, China}
			\IEEEcompsocitemizethanks{\IEEEcompsocthanksitem $\diamond$ Jiajun Liu is the primary contact author.}
		}

	%%%\vspace{-1cm}
	
	\pagenumbering{arabic}
	%\pagenumbering{gobble}
	\setcounter{page}{1}
	
	\maketitle
	
	\begin{abstract}
		We have witnessed the discovery of many techniques for network representation learning in recent years, ranging from encoding the context in random walks to embedding the lower order connections, to finding latent space representations with auto-encoders. However, existing techniques are looking mostly into the local structures in a network, while higher-level properties such as global community structures are often neglected. We propose a novel network representations learning model framework called RUM (network Representation learning throUgh Multi-level structural information preservation). In RUM, we incorporate three essential aspects of a node that capture a network's characteristics in multiple levels: a node's  affiliated local triads, its neighborhood relationships, and its global community affiliations. Therefore the framework explicitly and comprehensively preserves the structural information of a network, extending the encoding process both to the local end of the structural information spectrum and to the global end. The framework is also flexible enough to take various community discovery algorithms as its preprocessor. Empirical results show that the representations learned by RUM have demonstrated substantial performance advantages in real-life tasks.
	\end{abstract}

	\section{Introduction}
	\label{sec:intro}
		Through time and social progress, networks become ubiquitous in our daily life, representing one of the most important form of data nowadays. A member of the modern society deals with various kinds of networks on a daily basis, e.g. colleague networks, family networks, even mobile phone networks. Not only humans, but every group of entities that develop links between one another can be considered a network. 
		
		Network research has therefore played a vital role in many real-life applications. For example, two Facebook users that share many followers and followees as well as group memberships tend to have similar interests. Such property shows great potential on applications such as advertising, friend recommendation, influence and event propagation analysis etc. Many of these applications rely on the effective representation of each node, that is, a distinctive feature form that captures the structural and semantic information of networks. A common form of network representation is to use a distributed and dense vector to represent each node, and the process of generating such vectors is called network representation learning or network embedding. It is widely proved effective in node classification, clustering, link prediction, visualization and other applications.

		Researcher have developed a series of methods that demonstrate competitive performances. DeepWalk \cite{Perozzi:2014:DOL:2623330.2623732} uses random walks and the Skip-gram model promoted by Word2Vec \cite{NIPS2013_5021} to learn the embeddings. Some researchers proved that random walk-baed embedding generation is equivalent to with matrix factorization \cite{levy2014neural}. The essence of network embedding is to determine what kind of network proximities is effective for the preservation of structural and semantic information of a network. In light of discovering meaningful proximities, \cite{Tang:2015:LLI:2736277.2741093} defines the first-order and second-order proximities, focusing on capturing local information of a network. While some higher-order proximity such as neighborhood proximity \cite{grover2016node2vec} is proposed, communities as a form of global structural information, is often overlooked.  ComEmbed \cite{DBLP:journals/corr/ZhengCCCC16} first introduced the community structure to network embedding and achieve community embedding. Then Community-enhanced Network Representation Learning (CNRL) \cite{DBLP:journals/corr/TuWZLS16} model and Community Preserving Network Embedding (CPNE) \cite{DBLP:conf/aaai/WangCWP0Y17} model are presented to incorporate the community structure and node local proximity when embedding the network. However, we argue that most of the existing work either focuses on the local end in the network information spectrum, or emphasizes on the global end mostly, and most of them does not distinguish the tightness of communities. A embedding method that is able to comprehensively integrate multiple levels of structural information is of urgent need.

		An analogy to this perspective is that a person's engagement with the society exists in multiple levels, ranging from close friend circles, alumni networks, employer organization, unions, to something of larger-scale such as residing cities and even nations. A person can be affiliated to many organizations in the same level at the same time. Encouraged by this intuition, we design a network embedding model framework that preserves community structures at multiple scales covering the entire structural information spectrum. 
		
		To distinguish the levels of community structures and extract information from them accordingly, we define two novel proximities, the triadic proximity and the global community proximity.  The two proximities measure a node's very close encounters with others and its position in the entire network perspectively. The triadic proximity measures the proximity of small and dense community structure and assumes that the nodes in a dense community are more likely to be closely related. The global community structure provides effective and rich information on the network level. As a supplement, the neighborhood proximity grasps the lower-to-middle-level information. Rather than finding community and embed the network simultaneously like other methods, we devise a flexible framework that can use community structures discovered by any community detection algorithms, and then carry on network embedding, obtaining more discriminative network representations. 
		
				\begin{figure*}[htp]
			%%\vspace{-0.3cm}
			\hspace{0.5cm}
			%\centering
							%\hspace{-1.4cm}
				\includegraphics[scale = 0.35]{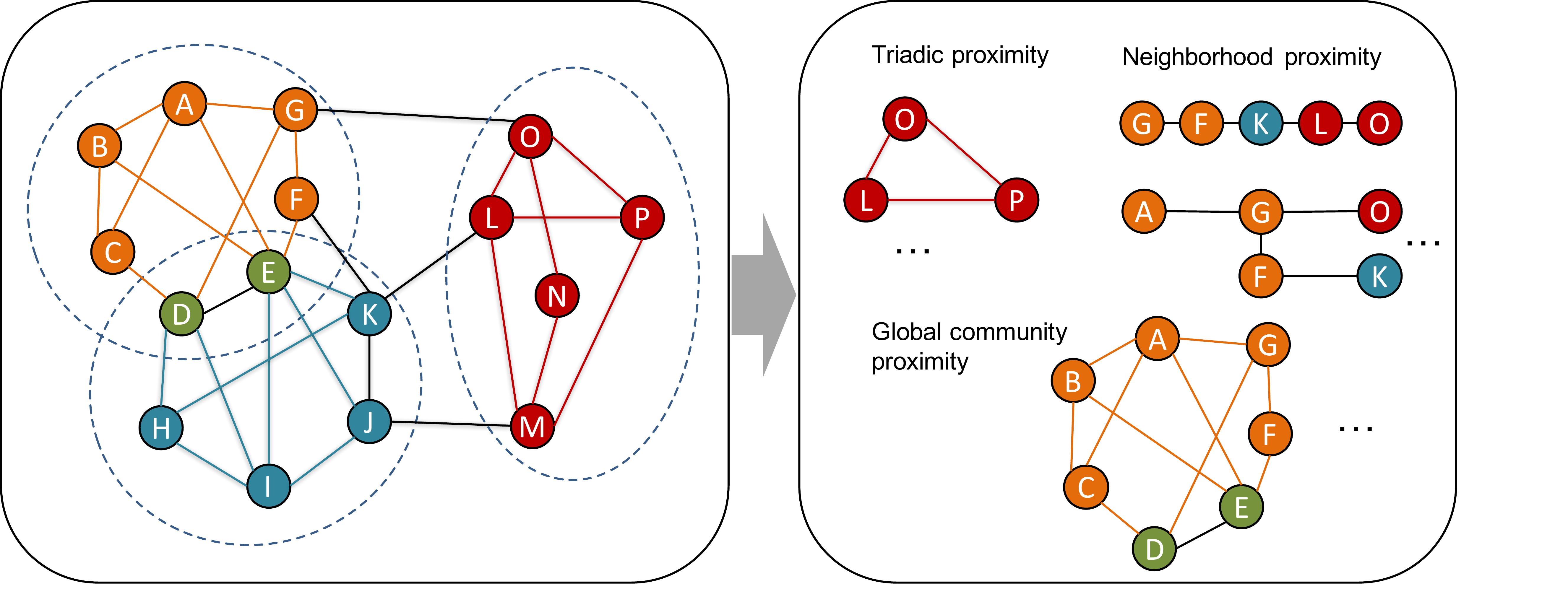}        \label{fig:intro}
			\caption{Conceptual visualization of multi-level structural information preservation via triadic proximity, global community proximity and neighborhood proximity. }
			\label{fig:intro}
			%%%\vspace{-0.6cm}
		\end{figure*}
		
		A conceptual visualization of the discussed multi-level structural information is presented in Figure \ref{fig:intro}. A node in a network often resides in structures of different scales, and together these structures reflect the node's  position in a network from different perspectives. A triad means a triplet of nodes in which all members are interconnected, mimicking the  smallest  multi-party close social relationship defined by Georg Simmel \cite{Leck:2000aa}. An individual often needs to make an effort to cope with more complex social relationships as a member of many larger groups (such as workplace circles), and therefore must hold tight on their affiliated smaller groups (family and close friends). On the other hand, global communities are larger groups that exhibit relatively loose connections among members. Finally, neighborhood proximity captures generalized direct and indirect connections through one or more direct links. Though neighborhood proximity may reflect triadic and larger communities to a certain degree, its ability to do so is substantially constrained by sampling strategy and computational costs. We argue that the explicit integration of these three aspects leads to a more comprehensive and effective way to preserve structural information at multiple levels.

		Our contribution is threefold:
		\begin{itemize}
		\item We propose two novel kinds of proximity measurements among network nodes, namely the triadic proximity and the global community proximity, extending the network embedding techniques to both the fine-grained level and the omniscient end for the entire network.
		\item We pioneer the integration of structural information at three levels in a network for representation learning in the proposed model framework. We exploit triadic community, global community and neighborhoods simultaneously, and devise a network embedding framework that preserves network structures effectively.  We call the framework RUM (network Representation learning throUgh Multi-level structural information preservation).
		\item We conduct extensive experiments and thoroughly evaluate the proposed framework, and demonstrate its advantages over existing methods.
		\end{itemize}

		The remainder of this paper is organized as follows: we present preliminary definitions and concepts in Section \ref{sec:pre}, and then elaborate on the technical details of RUM in Section \ref{sec:model}. Then we  conduct extensive experiments and analyze empirical results in Section \ref{sec:exp}, followed by a comprehensive survey of existing literature in Section \ref{sec:rw}. Finally we conclude our work and discuss future developments of the proposed method in Section \ref{sec:conc}.
	
	%%\vspace{-0.1cm}

	\section{Preliminaries}
	\label{sec:pre}
	%%\vspace{-0.1cm}

	In this section, we present the concepts, definitions and symbols used throughout the paper and introduce the problem we study in the paper. First we introduce the definition of information network:
	
	\definition[\textbf{Information Network}] An information network is composed of a set of nodes, and a set of edges between pairs of nodes. We use $\mathcal{G}(\mathcal{V},\mathcal{E})$ to denote a network, where $\mathcal{V} = \{v_1,v_2,...,v_n\}$ is the set of all nodes and $n$ is the number of nodes in a network. An edge $e_{ij}=(v_i,v_j)$ is used to denote a connection between nodes $v_i$ and $v_j$, and $\mathcal{E} =\{e_{ij}\}$ is the set of all edges in a network $\mathcal{G}$.
	
	 In this paper, we mainly focus on undirected networks, but we see no difficulties in the application of RUM to directed networks too.
	
	A network is often represented by an adjacency matrix $A$, where each entry $A_{ij}$ indicates if an edge between the $i^{th}$ and the $j^{th}$ nodes exists. A node $v_i$ in the network could be hence represented as an adjacency vector $A_i$. Adjacency information captures the most essential yet shallow information of a network, and it shows numerous limitations naturally: adjacency matrix does not explicitly show any information besides direct connections, and the lack of higher-order or higher-level information in adjacency matrix implies that this feature form is not descriptive enough; similar to some one-hot feature representations, an adjacency matrix's dimensions are determined by the number of nodes, and is often sparse, resulting severe space deficiencies. Motivated by the need of more compact and informative feature representations for networks, embedding methods are studied in recent years. For example, Word2Vec learns the embedding vectors of words and obtains the distributed representations so that similar words have similar representations. Network embedding is defined as follows:
	
	\definition[\textbf{Network Embedding}]  Network embedding is the process of learning dense and relatively low-dimensional distributed continuous vectors for each node in a network to capture distinctive and descriptive hidden features of the nodes. In other words, network embedding learning aims to learn a function $f$ which transforms each node to a point in a $d$-dimensional ($d << n$) latent feature space where ``similar'' nodes have close distances.
	
	\definition[\textbf{Network Proximity}] Network proximity the measurement of the similarity between two nodes. Nodes with a shared edge or with many shared neighbors are considered more similar in a network. These cases are defined as \textit{first-order proximity} and \textit{second-order proximity} respectively. 
	
	The design and preservation of proximity in the embedding process is vital. A majority of the literature seeks to preserve one or both proximities effectively. For example, DeepWalk \cite{Perozzi:2014:DOL:2623330.2623732} uses random walks to generate sample sequences and considers nodes in the same window in a walk are similar to each other. LINE \cite{Tang:2015:LLI:2736277.2741093} chooses to minimize the KL-divergence of the empirical probability and embedding probability based on first-order and second-order proximity. Effectively these methods capture only the local proximity and neglects higher-order or even global proximity, which is also significant information for nodes. To overcome this weakness, we propose the concept of \textbf{Global Community Proximity} to preserve structural information in a global level. Moreover, to reflect the tight local structures more effectively, we introduce \textbf{Triadic Proximity}. And the generally used local co-occurrence relationships is defined as \textbf{Neighborhood proximity} in our work. Together, the three proximities form a full spectrum of structural information that we see to preserve in the representation learning process.
	
	\eat{
		and define different proximity in accordance with different community structure. 
		
		In this paper, we aim to devise a better network embedding method that captures the proximity between nodes in multiple levels and multiple facets. To differentiate these mutually-complementary aspects,  we introduce the concepts of  \textbf{Triadic Proximity}, \textbf{Global Community Proximity}, and \textbf{Direct link proximity}, .
	}

	\eat{
		To overcome the problem, aspects that concern high-order proximities are promoted, and community structures are getting attention in recent research,  to achieve better embedding performance. Some researches consider community structure and combine community embedding with node embedding to ensure the community vector is similar with the node vector. But there is no precise community proximity definition put forward to better solve the problem. Otherwise, as there is no ground truth community result in network, they usually make use of iteration methods to determine affiliation between nodes and communities and generate embedding vector simultaneously, which leads to poor community detection. To solve this problem, we introduce precise definition of community proximity and define different proximity in accordance with different community structure.}
		
	First we propose \textbf{Triadic Proximity}. Intuitively, similar entities have a greater probability to consociate with the same community, hence nodes share many communities are more similar. However, not all communities reflect the same level of closeness. For example, a \textit{clique} is the densest structure in a network, which means that every node in the community has an edge with every other node in this community. Then for a \textit{k-clique} which has $k$ nodes, an equivalent definition is that every node connects to every other node with k-1 \textit{triads}, where a triad is defined as a clique with three inter-connected nodes. As the minimum form of a clique, triads constitute a key element to greater cliques. It is natural to consider the number of triads an important measurement for how tight a community is. For example, if there is a community with $k$ nodes, there are at most unique $\frac{k(k-1)(k-2)}{6}$ triads. Naturally the fraction of the actual number of triads to the theoretically upper-bound can be used as a metric of the tightness of the community. The more triads two nodes share, the tighter there connection is, and therefore they are more likely to be similar. Based on this intuition, we propose to use Triadic Proximity as a measurement of the tightness of the shared communities between two nodes:
	\definition[\textbf{Triadic Proximity}]The triadic proximity of nodes $v_i$ and $v_j$ is define as:
	\begin{equation}
		|\mathcal{T}(v_i,v_j)|.
	\end{equation}
	, where $\mathcal{T}(v_i,v_j)$ is the set of shared triads for nodes $v_i$ and $v_j$.
	
	Now we re-visit the concept of \textbf{Network Community}. In reality,  members in a greater group often form subgroups based on hobbies, interests, education experience, geographical regions, etc. The citation network of scientific papers has many communities that represent different research areas. Furthermore, in the real world, a person joins in different communities for different reasons, thus one can be affiliated to multiple communities and communities are therefore overlapped. Naturally, we consider network community in the embedding process. Assuming that every node in a network is affiliated to zero or more communities, and the communities can be overlapped, we use $\mathcal{C} = \{\mathcal{C}_1,\mathcal{C}_2,...,\mathcal{C}_m\}$ to denote the community set for a network. Then we use the affiliations matrix $R\in\mathcal{R}^{n\times m}$ to indicate the affiliation relationships between nodes and communities, where $R_{ij} = 1$ if node $v_i$ is a member of community $\mathcal{C}_j$. 

	Though triadic proximity captures tight local circles in a network, there still exists more structural information in the global scale of a network to be preserved.  That is, even two nodes may not share any triads, the fact that they appear in the same community indicates a certain level of similarity. To reflect this aspect, we propose global community proximity: 
	\definition[ \textbf{Global Community Proximity}] The global community proximity of nodes $v_i$ and $v_j$ is define as:
	\begin{equation}
		|\mathcal{C}(v_i,v_j)|.
	\end{equation}
	, where $\mathcal{C}(v_i,v_j)$ is the set of shared communities for nodes $v_i$ and $v_j$.
	The community structures can be discovered by many existing methods such as Affinity Propagation or BIGCLAM \cite{DBLP:conf/wsdm/YangL13}.
	
	In addition, not sharing any kind of communities does not indicate that two nodes are completely unrelated. The neighborhood proximity is also important to consider in network embedding. Similar to many existing methods, we consider co-occurrences in a randomly generated path on the network an indicator of the neighborhood proximity:
	\definition[\textbf{Neighborhood proximity}] The neighborhood proximity $w_{ij}=1$ if there exists co-occurrence for nodes $v_i$ and $v_j$ in a random walk, $0$ otherwise. 
	
	In the next section we will give an elaborated description for the proposed model framework, including the derivation and use of these proximities as well as the formulation of objectives and their optimization processes.

	\section{The Model Framework}
	\label{sec:model}
	The key motivation of the proposed model is to incorporate various levels of structural information of a network to improve embedding quality. Shifting from the local end to the global end of the spectrum in a network's structural information, we consider the three types of proximities: the triadic proximity, the global community proximity, and the neighborhood proximity. Here we formulate the objective functions to preserve these proximities.
	
	\eat{To model these proximities, for each pair of nodes $(v_i,v_j)$, first we use $u_i$\yu{$\overrightarrow{u_i}$} and $u_j$\yu{$\overrightarrow{u_j}$} to represent their embedding vector, and define their base similarity as:
		\begin{equation}
			\mathcal{P}(v_i,v_j) = \frac{1}{1+\exp(-\overrightarrow{u_i}^T \cdot \overrightarrow{u_j})}
		\end{equation}
	}
	
	\subsection{Triadic Proximity}
	Triadic proximity measures the number of triads two nodes share. Higher triadic proximity not only indicates that two nodes co-exist in tight communities, but further reflects their shared interests and social circles could be many. To model the triadic proximity, for each pair of nodes $(v_i,v_j)$, first we use $u_i$ and $u_j$ to represent their potential embedding vectors respectively, and define their similarity as:
	\begin{equation}
		\mathcal{P}_1(v_i,v_j) = \sigma(\overrightarrow{u_i}^T \cdot \overrightarrow{u_j}),
		\label{eqn:p1}
	\end{equation}
	where $\sigma(x)=\frac{1}{1+\exp(- x)}$ is the sigmoid function.
	
	We know the empirical triadic proximity is the number of triads $v_i$ and $v_j$ share, which can be represented in a triadic proximity matrix $T$. We can calculate T as:
	\begin{equation}
		T=A\odot A^2.
	\end{equation}
	,where $ \odot$ is the element-wise product operator and $A$ is the adjacency matrix, and the vector $A_i$ indicates the nodes which $v_i$ can arrive in one step. Similarly, vector $A^2_i$ measures the number of paths through which $v_i$ can arrive at other nodes in two steps. The element-wise product of $A$ and $A^2$ therefore estimates the number of paths through which $v_i$ arrives at $v_j$ in two steps while $v_i$ connects to $v_j$ directly, hence forming a triad. To preserve this triadic proximity, we require that $\mathcal{P}_1$ converge to $T$. Therefore, a straightforward method to implement this idea is to minimize the distance between these two distributions, as: 
	\begin{equation}
		\text{minimize} ~ d({T_{ij}},\mathcal{P}_1(v_i,v_j))
	\end{equation} 
 	where $d(,)$ is a function to measure the distance between two distributions. We then use the KL-divergence to measure the distance between them. Omitting some constants, our final objective should include the following objective function:
	\begin{equation}
		O_1= - \sum_{i,j\in\mathcal{V}}{T_{ij} \log \mathcal{P}_1(v_i,v_j)}.
	\end{equation}
	
	\subsection{Global Community Proximity}
	Global community proximity measures the similarity of two nodes by considering their co-occurrences in the same communities of large-scales. For a given network with community structures unknown, the first step is to detect its global communities . As one may take part in multiple organizations for different reasons in real life, we naturally assume the communities  are overlapping. Our framework support the use of different community detection methods. For example, in BIGCLAM \cite{DBLP:conf/wsdm/YangL13}, the algorithm first initializes a node community membership with a bipartite affiliation network that measures the likelihood of whether a node belong to a community. Then it assumes that one node has greater likelihood to connect to other nodes which share more common communities with it. Based on this intuition, the algorithm uses the matrix $F=\{F_{v_ic_j}\}$ to measure the likelihood of the node $v_i$ belong to the community $c_j$. And it then measures the edge probability between node $v_i$ and node $v_j$ as:
	\begin{equation}
		\mathcal{P}_{e_{ij}}=1-\exp(-\sum_{c_k\in \mathcal{C}} { F_{v_i c_k} F_{v_j c_k} }),
	\end{equation}
	where $\mathcal{C}=\{c_1,c_2,...,c_m\}$ is the set of all detected communities.
	
	Using $F\in R^{n\times m}$, where n is number of nodes and m is the number of communities, we can then generate a new network $\mathcal{G}'(\mathcal{V}',\mathcal{E}')$ and we expect $\mathcal{G}'$ to be similar to the original network $G(\mathcal{V},\mathcal{E})$. We use the loss function $D$ to estimate the distance between $\mathcal{G}$ and $\mathcal{G}'$ and also use $f(\cdot) = 1-\exp(\cdot)$ as the activate function:
	\begin{equation}
		\hat{F}=arg\min_FD(A, f(FF^T)).
	\end{equation}
	
	BIGCLAM revises the NMF methods and uses matrix factorization to solve this problem. By relaxing the objective function, we obtain the equivalent definition, which means that maximizing the likelihood $L(F) = \log P(\mathcal{G}|F)$ of the underlying $\mathcal{G}$:
	\begin{equation}
		\hat{F}=arg\min_F-L(F)
	\end{equation}
	where
	\begin{equation}
		L(F) = \sum_{v_i,v_j\in \mathcal{E}} \log(1-\exp(-F_{v_i}F_{v_j}^T)) - \sum_{v_i,v_j\notin \mathcal{E}} F_{v_i}F_{v_j}^T.
	\end{equation}
	By optimizing this objective we obtain the communities.
	
	After finding the communities in the network, we use the affiliation relationships to force that nodes in the same community have similar embeddings. To preserve such global community proximity, we consider the nodes in the same community $c$ as the context for each other. Besides, as a group nodes, a community is also considered as a entity, and we assign a $d$-dimension vector $\overrightarrow{c_i}$ as its feature vector. Then we assume that a pair of node and community has a certain degree of relationship. Hence we measure the strength of relationships between community $c_i$ and node $v_j$ by the conditional probability of having nodes $v_j$ when given the community and hence define the conditional probability of having nodes $v_j$ when given the community $c_i$:
	\begin{equation}
		\mathcal{P}_2(v_j|c_i) = \frac{ \exp(\overrightarrow{u_j'}^T \cdot \overrightarrow{c_i}) }{ \sum_{k=1}^n \exp(\overrightarrow{u_k'}^T \cdot \overrightarrow{c_i}) }.
	\end{equation}
	 It is natural that any node $v_j$ that belongs to a community tend to have a similar embedding vector to $c_i$. Therefore we use the softmax function to construct the conditional probability of the embedding vectors given its affiliated community embedding. On the other hand we have obtained the community affiliations as a matrix $R$ from the community detection algorithm, now we force $\mathcal{P}_2(v_j|c_i)$ to satisfy $R$. Intuitively, when given a community $c_i$, the ``impact'' of a node $v_j$ in the community can be measured by the numbers of the $v_j$'s neighbors which belong to $c_i$ too. The more friends of $v_j$ in $c_i$, the more committed $v_j$ is to $c_i$. To measure the degree of involvement, we define the matrix $S\in \mathcal{R}^{n\times m}$ as $S=AR$, where $S_{ji}$ estimates the number of neighbors of $v_j$ that belong to community $c_i$. Given a community $c_i$, we calculate the number of neighbors in this community for each node and obtain the normalized wight $sum^n_{k=1} S_{ki}$ . Then we can estimate the empirical conditional probability which estimates the likelihood of $v_j$ belonging to $c_i$, as:
	
		\begin{equation}
		\label{eqn:ph2}
		\hat{\mathcal{P}_2}(v_j|c_i) = \frac{ S_{ji}}{\sum^n_{k=1} S_{ki}},
		\end{equation}

	A higher probability indicates that the node is more likely one of the members in community $c_i$. The empirical condition probability and embedding conditional probability essentially measure the identical probability that one node belong to the community from two angles. For a quality embedding, the results should accurately measure the strength of relationship between node and community similar to the empirical inference and the two distributions should be converge. Thus we use the conditional KL-divergence to measure the distance between two distributions and minimize their KL-divergence. By omitting some constants, we arrive at:
		\begin{equation}
		O_2 = - \sum_{v_j\in \mathcal{V}, c_i\in \mathcal{C}} S_{ji} \log \mathcal{P}_2(v_j|c_i).
		\end{equation}		
	$O_2$ estimates the probability of nodes given a community. 
	On the other hand, we consider that given a node $v_j$, its community affiliations should be able to be inferred. Based on the community embedding and node embedding, we can obtain the conditional probability of each community given node $v_j$, as:
	\eat{\yu{On the other hand, we can talk about the relationship between nodes and communities from the point of view of the node. We consider that given a node $v_j$, its community affiliations should be able to be inducted. Similar to the conditional probability given the community $c_i$, we can obtain the conditional probability of each community given node base on the community embedding and node embedding, as:}}
		\begin{equation}
		\mathcal{P}_3(c_i|v_j) = \frac{ \exp(\overrightarrow{c_i'}^T \cdot \overrightarrow{u_j}) }{ \sum_{k=1}^m \exp(\overrightarrow{c_k'}^T \cdot \overrightarrow{u_j}) },
		\end{equation}
		where community $c_i \in \mathcal{C}$.
	\eat{\yu{It is straightforward to understand the equation based on the view that the nodes belong to the community $c_i$ should have a close relationship with the community, and the reaction on embedding space is that their embedding is similar. Based on this assumption, the nodes in the same community have similar embedding, which is similar to the community embedding simultaneously.}}
	
	Furthermore, similar to Equation \ref{eqn:ph2}, the relationships between $v_j$'s neighbors and each community can be incorporated into the objective. If a node has many neighbors belong to a community, the node itself is more likely belong to the community too. We again can estimate the empirical conditional probability for each community $c_i$ as:
	\begin{equation}
	\label{eqn:ph3}
	\hat{\mathcal{P}_3}(c_i|v_j) = \frac{ S_{ji}}{\sum^m_{k=1} S_{jk}}.
	\end{equation}
	Similarly, after defining the embedding conditional probability and the empirical probability, we can require the two probability distributions to be similar, hence we obtain the embedding objective function:
		\begin{equation}
		O_3 = - \sum_{v_j\in \mathcal{V}, c_i\in \mathcal{C}} S_{ji} \log \mathcal{P}_3(c_i|v_j).
		\end{equation}
	
	Objectives $O_2$ and $O_3$ simultaneously learn the community embeddings and node embeddings by forcing  similar communities and nodes to have similar embeddings. To integrate the objective functions, a new perspective can be introduced. The more communities two nodes share, the more similar they are, and therefore the more penalty they receive when their embeddings diverge. Here we introduce matrix $H = R R^T \in \mathcal{R}^{n\times n} $ with the assistance of the affiliation matrix $R$ obtained from community detection, where $H$ measures the empirical global community proximity and $H_{ij}$ indicates the number of communities which node $v_i$ and $v_j$ share. Similar to Equation \ref{eqn:p1}, we apply the KL-divergence to force the embedding probability distribution to approach the empirical global community proximity. Then we can preserve the global community by the following objective function: 
	
		\begin{equation}
		O_4 = - \sum_{i,j\in \mathcal{V}} H_{ij} \log \mathcal{P}(v_i|v_j),
		\end{equation}
		where $H = R R^T$. If one decides to use the community discovery results from other algorithms, one can simply replace $R$ to the affiliation matrix discovered by other methods. 
	
	\subsection{Neighborhood Proximity}
	In addition to the higher-order proximity which we discussed above, the neighborhood proximity holds vital information for the capturing of local structures which greatly complement triadic proximity. If two nodes are considered neighbors, their neighborhood proximity can be estimated by the empirical probability:
	\begin{equation}
	\label{eqn:ph4}
	\hat{\mathcal{P}_4}(v_i,v_j) = \frac{ W_{ji}}{\sum^n_{k=1} \sum^n_{s=1}W_{ks}},
	\end{equation}	
	where $W_ij=1$ if $v_i$ and $v_j$ are neighbors, and $0$ otherwise.
	Simultaneously, we can get the probability $\mathcal{P}_4(v_i,v_j)$ from the embedding vectors by applying dot product on $v_i$ and $v_j$, similar to Equation \ref{eqn:p1}. And by forcing $\mathcal{P}_4(v_i,v_j)$ to converge to $\hat{\mathcal{P}_4}(v_i,v_j)$ under KL-divergence, we have:
	\begin{equation}
	O_5 = - \sum_{(i,j)\in \mathcal{E}} W_{ij} \log \mathcal{P}_4(v_i,v_j).
	\end{equation}
	
	Random walks are commonly used to measure local neighborhood relationships, and node2vec \cite{grover2016node2vec} has discovered that by biasing the random walk probabilities towards BFS and DFS, the resulting walk paths seem to reveal local neighborhood relationships more effectively than uniformly performed random walks. Here we use a similar strategy: by performing random walks with two parameters to prioritize BFS or DFS, we generate a series of walk paths. Any pair of nodes $v_i$, $v_j$ that co-occurred in the same random path and in the same window (e.g a window of $5$ steps) would be considered neighbors and would therefore yield an $1$ for the corresponding $w_{ij}$.
	
	\subsection{The Final Objective}
	The effective representation of network nodes requires preservation of structural information at multiple levels in the network. By jointing considering the triadic proximity, the global community proximity and the neighborhood proximity, we have our final objective as:
	\begin{equation}
	O = - \sum_{i,j=1}^n (\alpha T_{ij} + \beta H_{ij} + W_{ij}) \log \mathcal{P}_4(v_i,v_j),
	\label{eqn:obj}
	\end{equation}
	where $\alpha$ and $\beta$ are two parameters to balance the weight of the three proximities. By minimizing the objective function, we can obtain the node embeddings. Details of the optimization process will be elaborated in the next sub-section.
	
	It can be proved that both DeepWalk and node2vec are special cases of RUM: by setting both $\alpha$ and $\beta$ to 0, the only information Equation \ref{eqn:obj} will encode is the neighborhood structure sampled by random walks, which makes RUM an equivalence to DeepWalk. If we follow the sampling strategy of controlling the random walks towards BFS or DFS, then RUM is effectively identical to node2vec. 
	
	\subsection{Model Optimization}
	Optimizing the objective function is computationally expensive, where time complexity reaches $n^3$. To improve speed and scalability, inspired by the negative sampling method the NCE training process \cite{Perozzi:2014:DOL:2623330.2623732}, we define an optimization process by stochastic gradient descent with NCE. Rather than calculating the summation over the entire set of nodes, we sample a set of positive and negative sample pairs to estimate the value of $\mathcal{P}(v_i,v_j )$.

	Positive samples include a pair of nodes $v_i$ and $v_j$ that are similar based on the triadic proximity, the global community proximity or the neighborhood proximity. And we use the embedding probability formula to measure the probability of co-occurrence of a positive pair:
	\begin{equation}
		\mathcal{P}_{pos}(v_i,v_j) = \sigma( \overrightarrow{u_i}^T \cdot \overrightarrow{u_j}).
	\end{equation}
	
	Now the probability of a negative sample can be computed as:
	\begin{eqnarray}
\mathcal{P}_{neg}(v_i,v_k) = 1 - \mathcal{P}_{pos}(v_i,v_k) = \sigma( -\overrightarrow{u_i}^T \cdot \overrightarrow{u_k}).
	\end{eqnarray}
	Each positive sample may have different strength of proximity, so we define the value of proximity as weight of the positive sample probability and set -1 as weight for the negative samples. Specifically, we minimize the following objective:
	
	\begin{eqnarray}
	\hat{O} &=& -\sum_{j=1}^{K_1} (\alpha T_{ij} + \beta  H_{ij} + W_{ij}) \log \sigma(\overrightarrow{u_i}^T \cdot \overrightarrow{u_j}) \\&-& \sum_{k=1}^{K_2} \log \sigma(-\overrightarrow{u_i}^T \cdot \overrightarrow{u_k}), \nonumber
	\end{eqnarray}
	
	Here a positive sample means there exists a linking edge, or at least one shared community, or at least one shared triad, for node pair $v_i$ and $v_j$. Otherwise the sample is considered a negative sample. $K_1$ and $K_2$ specify  the numbers of positive and negative samples perspectively. Consequently, the time complexity for a training iteration is reduced to $O(K_1+K_2)$.We improve the sampling strategy to obtain better embeddings by setting $\mathcal{P}_n(v)\propto d_v^{\frac{3}{4}}$ where $d_v$ is the degree of $v$ \cite{Perozzi:2014:DOL:2623330.2623732}. 

	We adopt stochastic gradient descent to optimize the final objective function. Therefore the gradient for each positive sample and negative sample to update is set as:
	\begin{eqnarray}
		\frac{\partial{\hat{O}_{pos}}}{\partial {\overrightarrow{u_i}}} &=& (\alpha T_{ij} + \beta H_{ij} + W_{ij}) \sigma(-\overrightarrow{u_i}^T \cdot \overrightarrow{u_j}) \overrightarrow{u_j}\\
		\frac{\partial{\hat{O}_{pos}}}{\partial {\overrightarrow{u_j}}} &=& (\alpha T_{ij} + \beta H_{ij} + W_{ij}) \sigma(-\overrightarrow{u_i}^T \cdot \overrightarrow{u_j}) \overrightarrow{u_i}\\
		\frac{\partial{\hat{O}_{neg}}}{\partial {\overrightarrow{u_i}}} &=& - \sigma(\overrightarrow{u_i}^T \cdot \overrightarrow{u_k}) \overrightarrow{u_k}\\
		\frac{\partial{\hat{O}_{neg}}}{\partial {\overrightarrow{u_k}}} &=& - \sigma(\overrightarrow{u_i}^T \cdot \overrightarrow{u_k}) \overrightarrow{u_i}
	\end{eqnarray}

	\section{Experiments}
	In this section we evaluate RUM extensively. First we perform two tasks, namely node classification and network reconstruction, as a quantitative evaluation, then we show a case study to analyze its embedding quality qualitatively.

			\begin{table}[htp]
%\centering
%\hspace{3cm}
\caption{Dataset Statistics}
%\hspace{-0.5cm}
\begin{tabular}{ |c|c|c|c|c| }
\hline & \textit{\textbf{email-Eu-core}} & \textit{\textbf{CoRA}} & \textit{\textbf{CiteSeer}} & \textit{\textbf{BlogCatalog}}\\
\hline $|\mathcal{V}|$ &  1,005&2,708&3,312& 10,312\\
\hline $|\mathcal{E}|$ &  25,571&57,884&4,715& 333,983\\
\hline $Avg.degree$ & 25.44 &21.38&1.42& 64.78\\
\hline $No.labels$ & 42 & 7&6& 39\\
\hline
\end{tabular}
\label{tbl:stat}
\end{table}

	\label{sec:exp}

	\begin{figure*}[htp]
			\vspace{0.3cm}
			\hspace{-0.5cm}
			\centering
			%\vspace{1cm}
			\subfigure[email-EU-core microF1]{
				%\hspace{-1.4cm}
				\includegraphics[scale = 0.42]{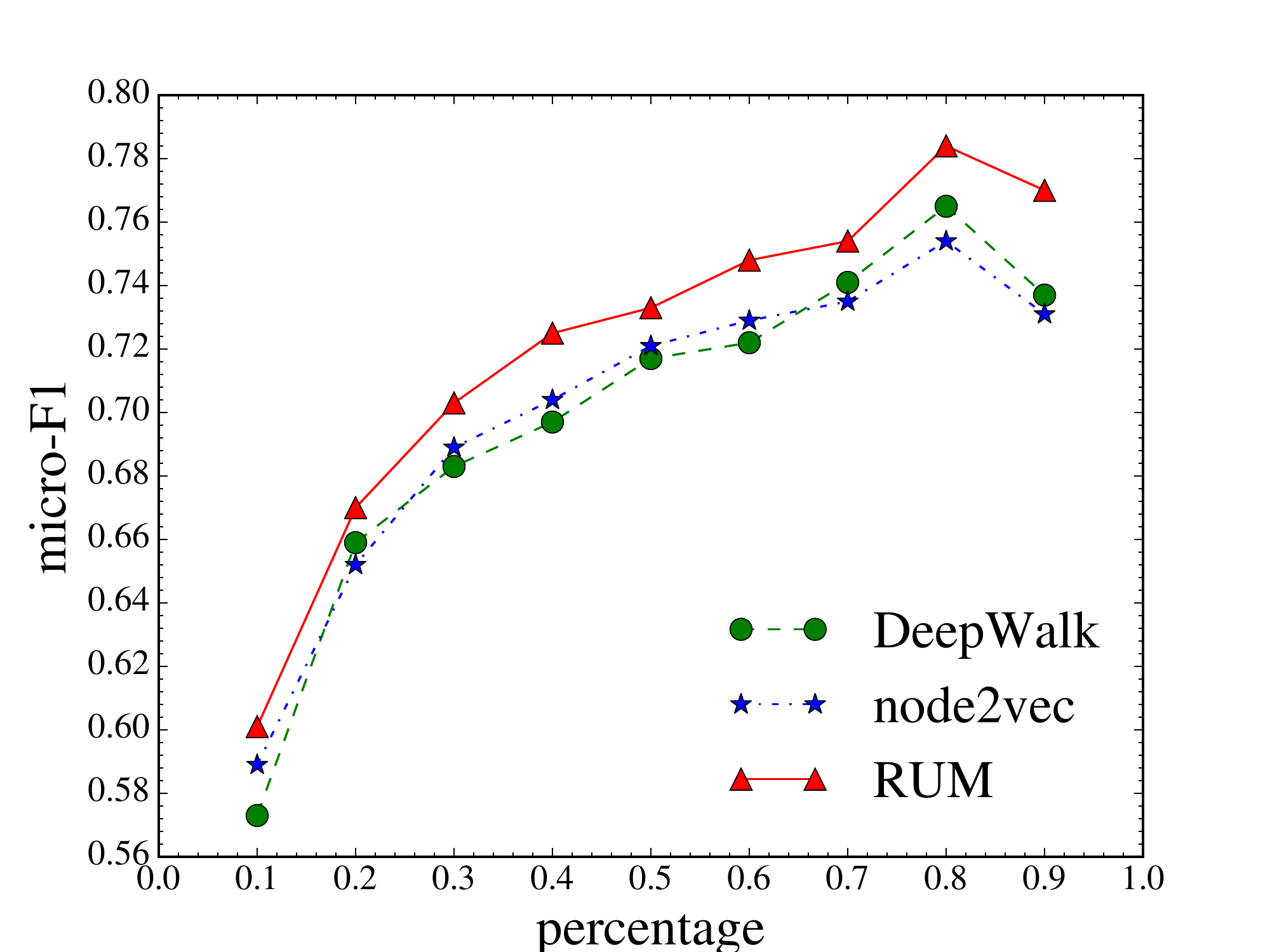}        \label{fig:nc1}
			}
			\subfigure[CoRA microF1]{
			%\hspace{-.6cm}
				\includegraphics[scale = 0.42]{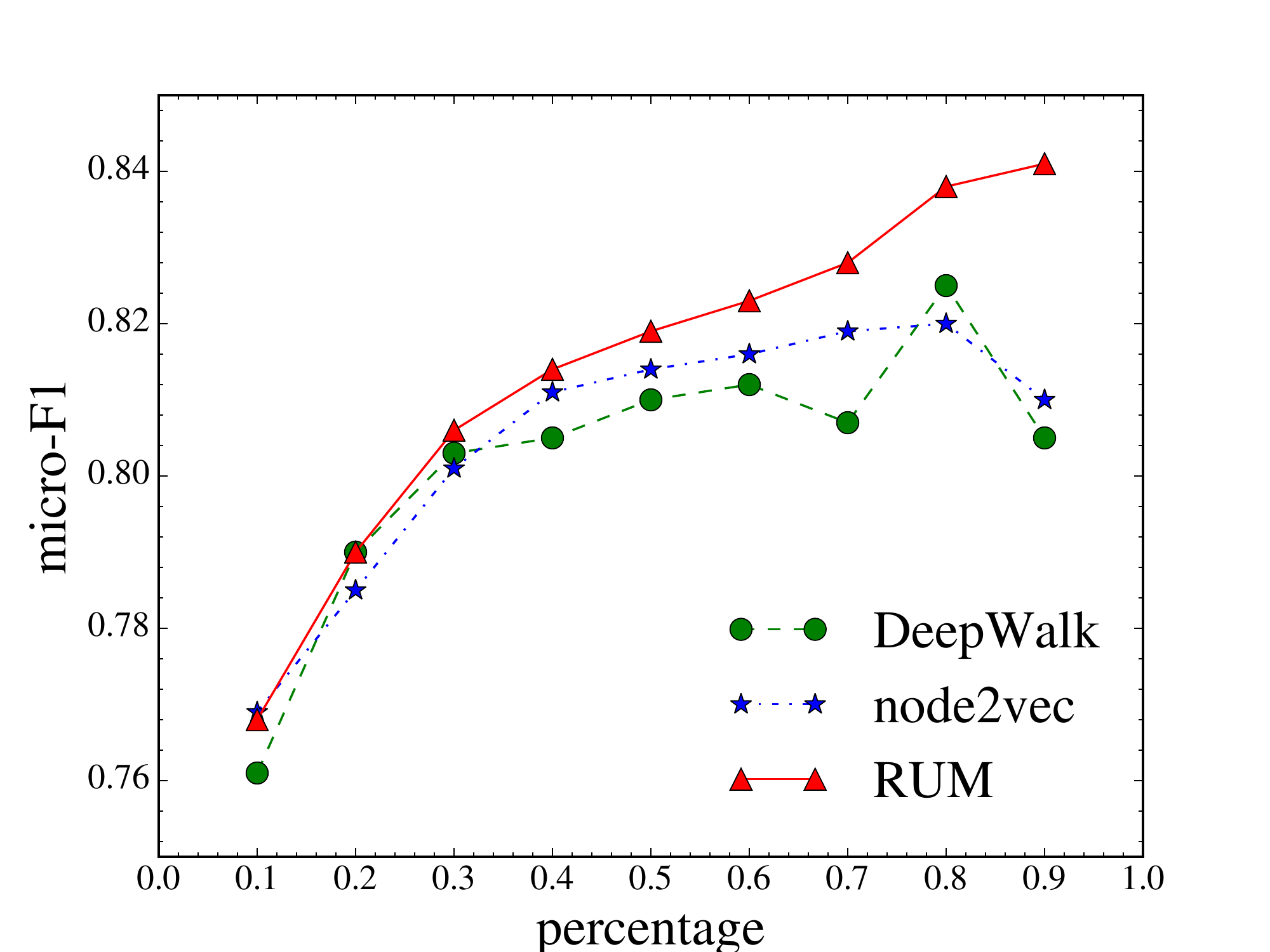}        \label{fig:nc3}
			}\\
			\subfigure[CiteSeer microF1]{
			% \hspace{-.6cm}
				\includegraphics[scale = 0.42]{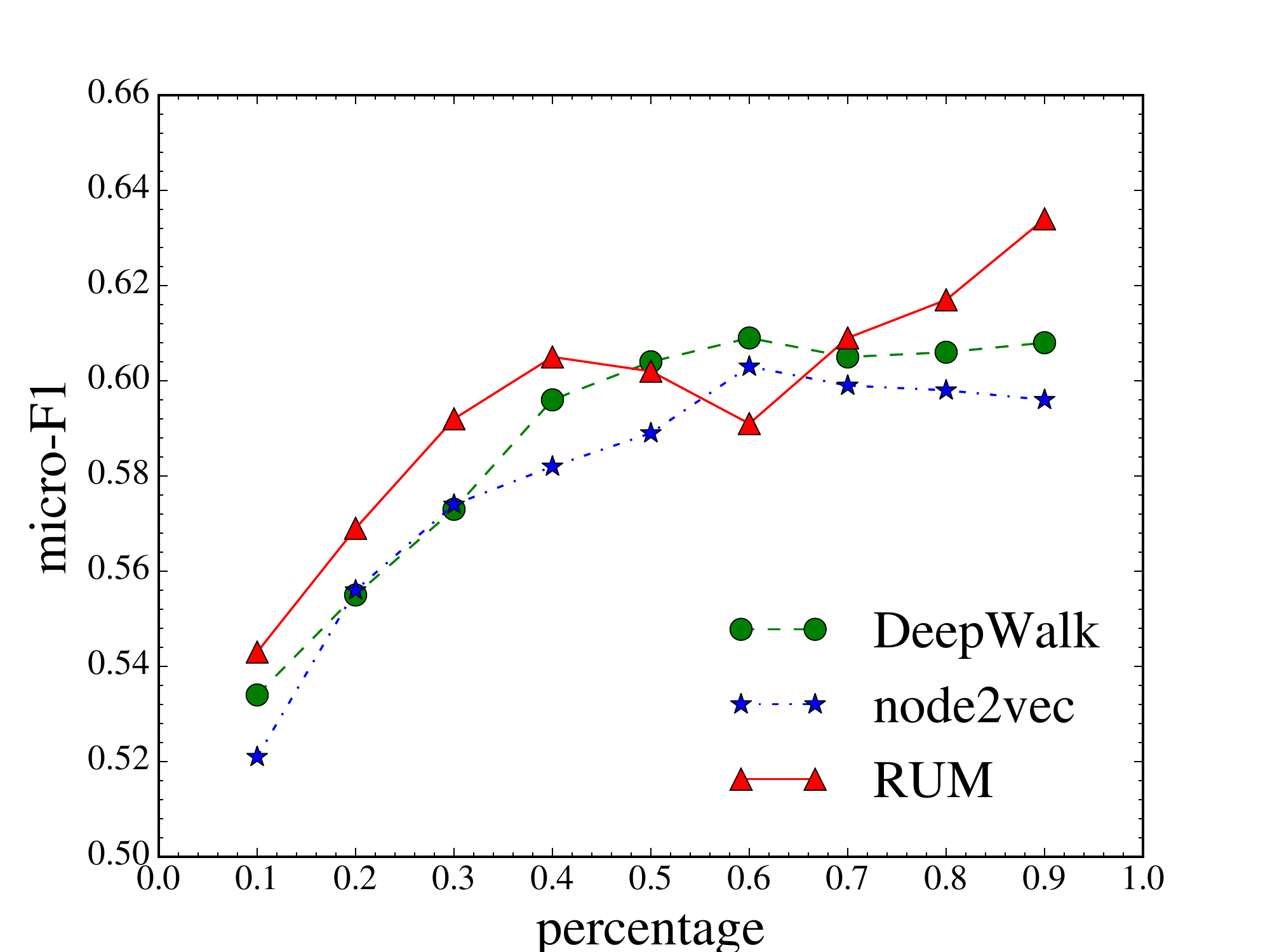}        \label{fig:nc5}
			}
			\vspace{0.3cm}
			\subfigure[email-EU-core macroF1]{
			 %\hspace{-1.4cm}
				\includegraphics[scale = 0.42]{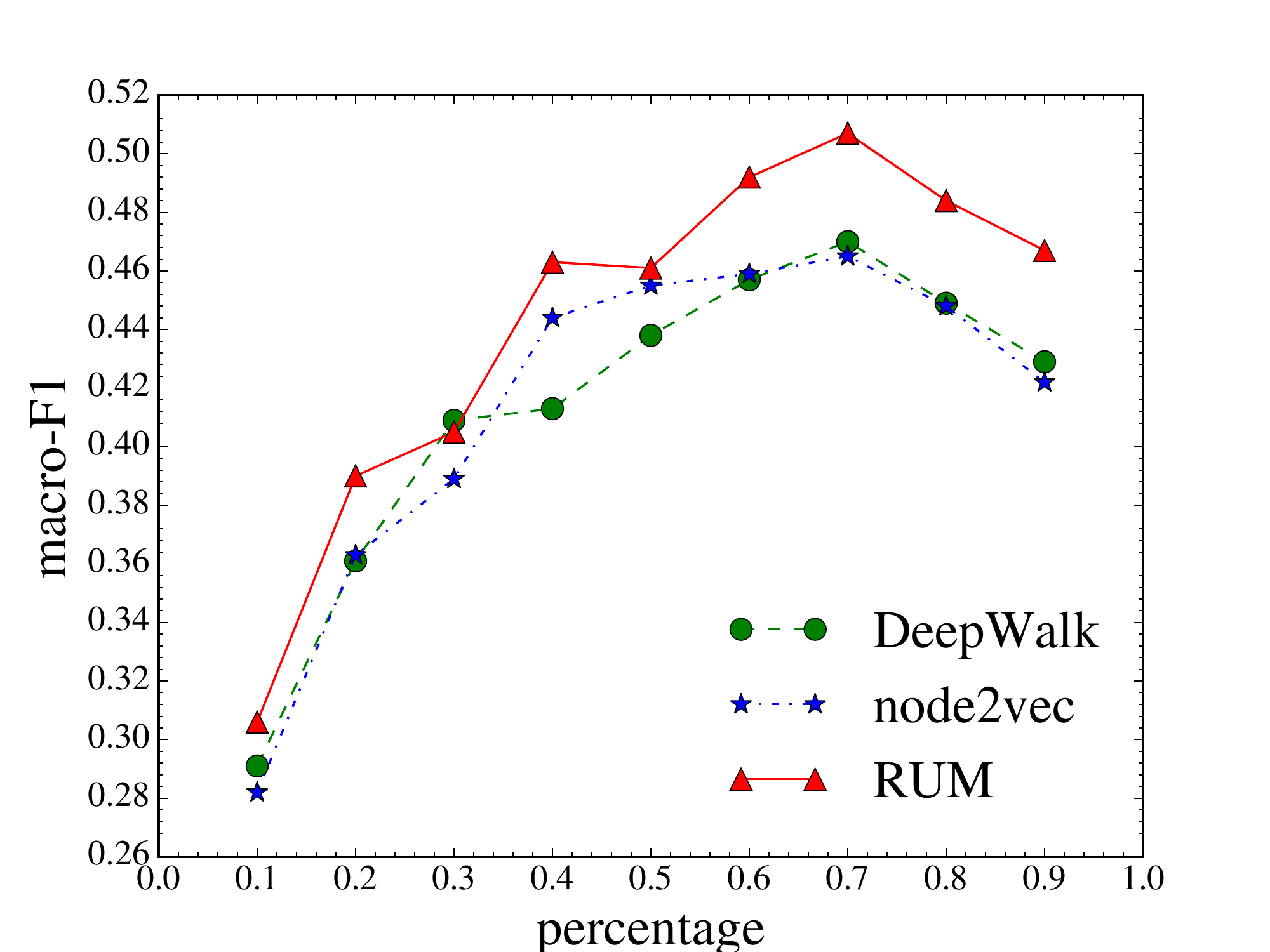}        \label{fig:nc2}
			}\\
			\subfigure[CoRA macroF1]{
			%	\hspace{-.6cm}
				\includegraphics[scale = 0.42]{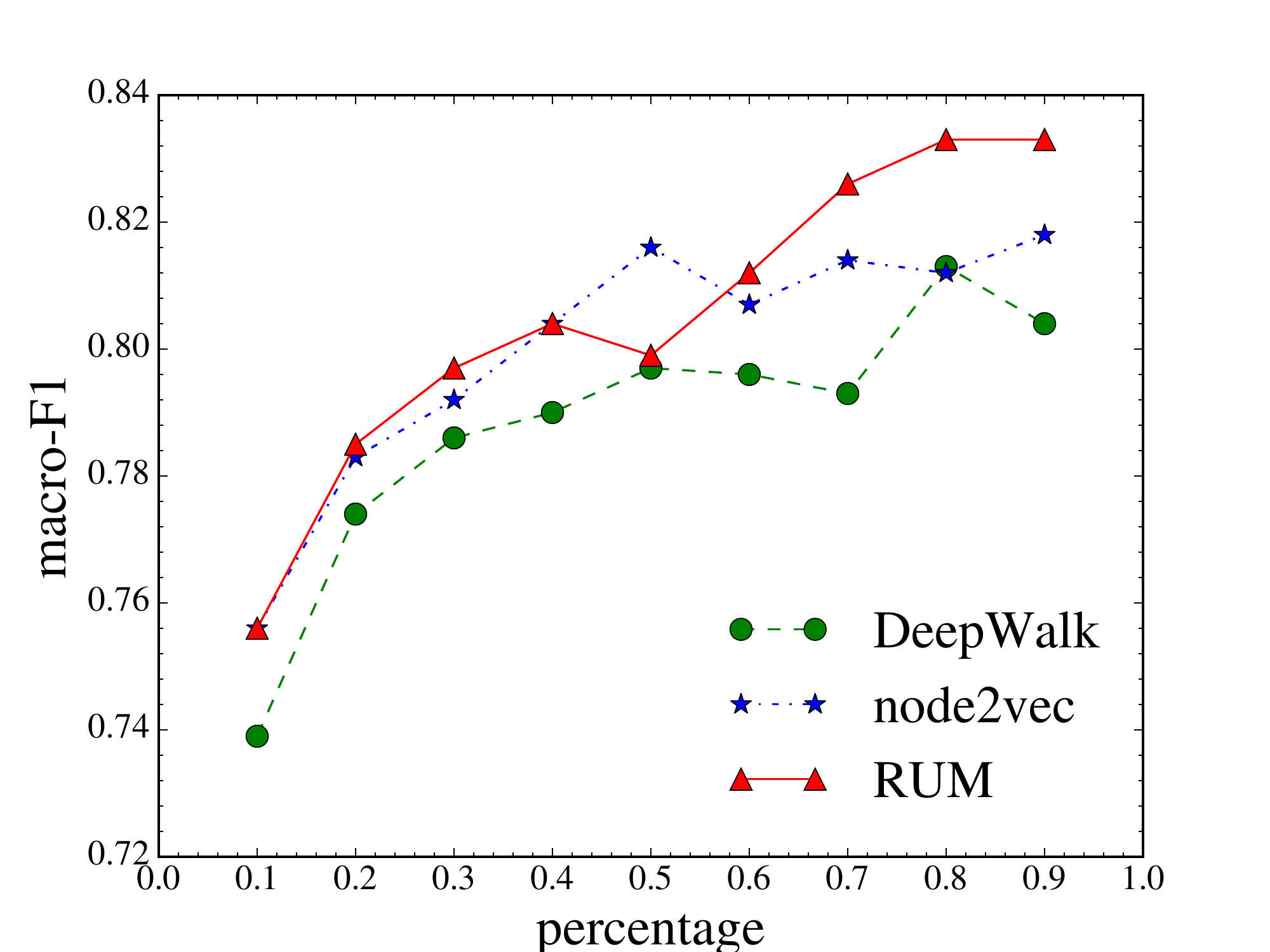}        \label{fig:nc4}
			}
			\subfigure[CiteSeer macroF1]{
			%	\hspace{-.6cm}
				\includegraphics[scale = 0.42]{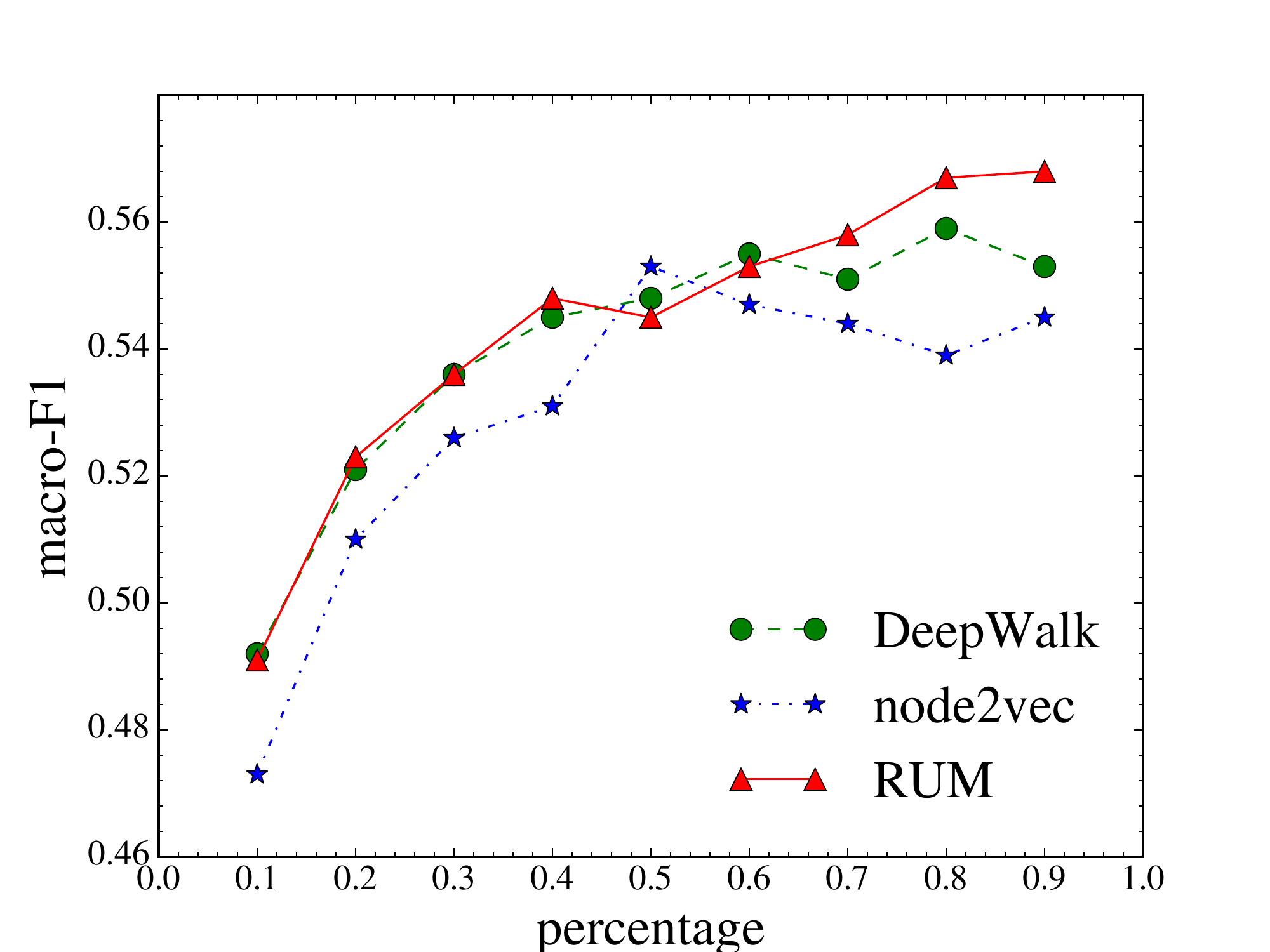}        \label{fig:nc6}
			}
			%%\vspace{-0.2cm}
			\caption{Node classification results on three datasets. } 
			\label{fig:nc}
			%%%\vspace{-0.6cm}
		\end{figure*}

	%%\vspace{-0.4cm}
	\subsection{Datasets}
	We use four datasets in our experiments:
	\begin{itemize}
	\item \textbf{\textit{email-EU-core}} \cite{Yin:2017:LHG:3097983.3098069}: this dataset consists of a network generated from email conversations from an European research institution. Each email user has a corresponding node in the network and if one user sends at least one email to another, they are considered connected in the network. The users belong to 41 departments in the organization, which correspond to 41 different labels. Email users from outside the institution are assigned under the ``others'' class.
	\item \textbf{\textit{CoRA}} \cite{DBLP:conf/ijcai/YangSLT17}: CoRA is a citation network that consists of 7 classes of papers, and each paper belongs to exactly one class. A citation from a paper to another forms a directed link between the two corresponding nodes.
	\item \textbf{\textit{CiteSeer}} \cite{DBLP:journals/corr/TuWZLS16}: CiteSeer is also a citation network that has 6 classes. 
	\item \textbf{\textit{BlogCatalog}} \cite{grover2016node2vec}: BlogCatalog is constructed from the social relationships from users on a blogger website. Based on a user's metadata, their interests are inferred and are used as labels. In total there are 39 different labels that correspond to user interests.
	\end{itemize}
	Statistics of the datasets are listed in Table \ref{tbl:stat}.

	%%\vspace{-0.1cm}
	
	%%\vspace{-0.4cm}
	\subsection{Evaluation Settings}
	\label{ssec:es}
		
	We compare RUM with three existing methods:
	\begin{itemize}
	\item \textbf{\textit{DeepWalk}} \cite{Perozzi:2014:DOL:2623330.2623732}: DeepWalk pioneered on using random walks to generate node co-occurrences and use Skip-gram to train embedding vectors. In this model, random walks are performed uniformly.
	\item \textbf{\textit{LINE}} \cite{Tang:2015:LLI:2736277.2741093}: LINE exploits the first and second-order proximities of the network explicitly and optimizes the embedding vectors towards a combination of the two. Specifically, we use ``LINE1'' to represent the half of the embedding vector that captures first-order proximity, and ``LINE2'' for the part that captures second-order proximity.
	
	\item \textbf{\textit{node2vec}} \cite{grover2016node2vec}: node2vec is also a random walk-based method which tweaks the sampling strategy of DeepWalk. Instead of uniform random walks, node2vec controls the random walk simulation process towards either BFS or DFS.
	\end{itemize}
	
	For each method (including RUM), we train embeddings of the same dimensionality ($d=128$), and use two tasks to evaluate the quality of the generated embedding vectors, namely node classification and network reconstruction. For both tasks we use the LogisticRegression implementation from Python Scikit-learn\footnote{\url{http://scikit-learn.org/stable/index.html}} with default parameters for the prediction. For each method, we conduct a series of cross-validations in which we gradually increase the training ratio from $10\%$ to $90\%$. To report empirical results, for each experiment, the cross-validation process is repeated five times for each. Finally the average results are reported.
	
	The evaluation metric we use is the F1-score, namely microF1 and macroF1.
	\begin{equation}
	F1 = \frac{2  (precision \times recall)}{precision + recall},
	\end{equation}
	where microF1 calculates the score globally by counting the total true positives, false negatives and false positives, and macroF1 calculates for each label, and finds their unweighted mean.
	
	\begin{table*}[htp]
%\vspace{0.3cm}
      \caption{Full Results on Node Classification}
      \label{tbl:comp}
      %\tiny
  \centering
    \begin{tabular}{c|ccc|ccc|ccc|ccc}
    \toprule
        \multirow{2}{*}{}   &  \multicolumn{3}{c}{email-EU-core}  & \multicolumn{3}{c}{CoRA} & \multicolumn{3}{c}{CiteSeer} & \multicolumn{3}{c}{BlogCatalog} \\
	MicroF1 & 10\%& 50\%& 90\ &10\%& 50\%& 90\%\ &10\%& 50\%& 90\%\ &10\%& 50\%& 90\% \\
	\midrule
	LINE1 & 0.52&0.73&0.743&0.529&0.658&0.669&0.378&0.455&0.472&0.344&0.4& {\textbf{\textcolor{blue}{0.412}}}\\
	LINE2 &0.419&0.608&0.653&0.363&0.425&0.453&0.235&0.283&0.294&0.294&0.347&0.371\\
	DeepWalk &0.573&0.717&0.737&0.761&0.810&0.805&0.534&0.604&0.608&0.344&0.376&0.381\\
	node2vec & 0.589&0.721&0.731&{\textbf{\textcolor{blue}{0.769}}}&0.0.814&0.81&0.521&0.589&0.596&0.331&0.359&0.362\\
	RUM &  {\textbf{\textcolor{blue}{0.601}}}& {\textbf{\textcolor{blue}{0.733}}}& {\textbf{\textcolor{blue}{0.77}}}& {\textbf{\textcolor{blue}{0.769}}}& {\textbf{\textcolor{blue}{0.819}}}& {\textbf{\textcolor{blue}{0.841}}}& {\textbf{\textcolor{blue}{0.543}}}& {\textbf{\textcolor{blue}{0.602}}}& {\textbf{\textcolor{blue}{0.634}}}& {\textbf{\textcolor{blue}{0.353}}}& {\textbf{\textcolor{blue}{0.382}}}&0.394\\
	          \bottomrule
    \end{tabular}%
\vspace{0.5cm}
 \begin{tabular}{c|ccc|ccc|ccc|ccc}
    \toprule
        \multirow{2}{*}{}   &  \multicolumn{3}{c}{email-EU-core}  & \multicolumn{3}{c}{CoRA} & \multicolumn{3}{c}{CiteSeer} & \multicolumn{3}{c}{BlogCatalog} \\
	MacroF1 & 10\%& 50\%& 90\ &10\%& 50\%& 90\% &10\%& 50\%& 90\%&10\%& 50\%& 90\% \\
	\midrule
	LINE1 & 0.319&	0.507	&0.43	&0.511&	0.649	&0.659	&0.344	&0.418	&0.43	&{\textbf{\textcolor{blue}{0.19}}}	&{\textbf{\textcolor{blue}{0.25}}}&	{\textbf{\textcolor{blue}{0.269}}}\\
	LINE2 &0.232	&0.406	&0.381	&0.272	&0.375	&0.403	&0.196	&0.237	&0.249	&0.145	&0.208	&0.226
	\\
	DeepWalk &0.291&0.438&0.429&0.739&0.797&0.804&0.492&0.548&0.553&0.171&0.21&0.224\\
	node2vec &0.282&0.455&0.422&0.756&0.816&0.818&0.473&0.553&0.545&0.153&0.181&0.194\\
	RUM &{\textbf{\textcolor{blue}{0.306}}}&{\textbf{\textcolor{blue}{0.461}}}&{\textbf{\textcolor{blue}{0.467}}}&{\textbf{\textcolor{blue}{0.756}}}&{\textbf{\textcolor{blue}{0.799}}}&{\textbf{\textcolor{blue}{0.833}}}&{\textbf{\textcolor{blue}{0.491}}}&{\textbf{\textcolor{blue}{0.545}}}&{\textbf{\textcolor{blue}{0.568}}}&0.179&0.212&0.229\\
	          \bottomrule
    \end{tabular}%
  \label{tab:res}%
%%\vspace{-0.5cm}
\end{table*}
	
	As a framework, RUM can take the communities discovery results from any algorithm. For simplicity, in the experiments, RUM uses the clustering results from conventional clustering methods (such as Affinity Propagation) as the preliminary global community information, and generates the $H$ matrix in Equation \ref{eqn:obj} accordingly. We also simply set $\alpha=\beta=1.0$, which means under such settings, the structural of low-level information of triads, neighbors and high-level information from global communities, are treated as equal parts. One can tweak these parameters and settings flexibly to fit their need. For example, one can use overlapping community detection algorithms and set a higher $\beta$ to emphasize the modeling of wider but looser social relationships; or one can set a higher $\alpha$ to force the model emphasize local structural information. Next we evaluate RUM with the node classification task.

	\subsection{Node Classification}
	In this task we perform  classification for the  network nodes. For citation networks CoRA and CiteSeer, label corresponds to a research area or conference. For email-EU-core, labels are the users departments. For each method and each dataset, the embeddings of the same dimensionality (d=128) are the input of the same LogisticRegression classifier, under default settings. We run five repetitions of cross-validation, with each repetition consisting of nine runs in which the train set ratio increases from 10\% to 90\%. The results are illustrated in Figure \ref{fig:nc}. Notice that because LINE's performance shows a large major disadvantage to the rest of the three methods, we report it later in the full result in Table \ref{tbl:comp}.
	
	We observe that by covering the full spectrum of structural information in the learning process, RUM is able to consistently show competitive performance on all datasets we evaluated under both microF1 and macroF1, and is able to outperform DeepWalk and node2vec by a substantial margin in most cases. Generally, RUM's advantage climbs as the training ratio increases, indicating that the model's learning ability scales with more information fed to it in the learning process, and it is able to capture the network structures more effectively.
	
	In email-EU-core, we notice a $5\%$ performance gain over DeepWalk and node2vec as training set ratio increases, under microF1 in Figure \ref{fig:nc1}. Similar trend is observed in email-EU-core under macroF1 in in Figure \ref{fig:nc4}. On CoRA, it is evident that RUM's performance gain widens steadily, and reaches 84\% microF1 at 90\% training set ratio, compared to those under 81\% for both DeepWalk and node2vec, in Figure \ref{fig:nc2}. On CiteSeer (Figure \ref{fig:nc3}), We also see that the performance gain is the highest (compared to the second-highest results) when training set ratio reaches 90\%, with RUM obtained 64\% microF1 and DeekWalk slightly lower than 61\%.
	
	A closer inspection on Figure \ref{fig:nc} indicates that the proposed RUM is rather robust on both the size and the density of the network. For example, CiteSeer exhibits sparse connections between nodes, where the average degree is less than two, while email-EU-core shows denser links in the network. RUM manages to perform well on both datasets. We argue that on sparser networks, RUM captures the structural information more through structural equivalence and high-level community information, while on denser networks, RUM is also able to capture the local tight circles and neighbors well too. Such flexibility introduced by the designed full coverage of multi-level structures is one of RUM's biggest advantages.
	
	The full results including BlogCatalog and LINE's empirical results are given in Table \ref{tbl:comp}. For each dataset and each training set ratio, we highlight the winning performance in blue text. The results show that RUM consistently achieves the best result under the same setting except for a few cases on BlogCatalog where LINE1 gives better results. However, notice that on all other dataset, LINE1's performance is far worse than RUM. For example, in email-EU-core, CoRA and CiteSeer, RUM outperforms LINE1 by up to 19\%,  44\% and 44\% perspectively. Overall RUM is the most competitive method in all methods evaluations.

	Next we present the results on network reconstruction.

	\subsection{Network Reconstruction}
	
	In this task we test the embedding quality by reconstructing a network from the learned embeddings. Specifically, first we train the network representation with RUM, and them use all the embeddings as nodes for a new network. By now all the edge information is absent. We then generate a new affinity matrix by calculating the Euclidean distance between each node from scratch. Finally for a node $v_i$ in the original network, we predict $|\{e_{ij}\}|$ links in the new network from the smallest $|\{e_{ij}\}|$ neighbors in the new affinity matrix, and evaluate the results with Mean Average Precision (MAP). The results are reported in Table \ref{tbl:rc}:
	
\begin{table}[htp]
\centering
\caption{MAP for Network Reconstruction on Four Datasets}	
\begin{tabular}{ |c|c|c|c|c| }
\hline & \textit{\textbf{email-Eu-core}} & \textit{\textbf{CoRA}} & \textit{\textbf{CiteSeer}} & \textit{\textbf{BlogCatalog}}\\
\hline LINE1 &  {\textbf{\textcolor{black}{0.666}}}&{\textbf{\textcolor{black}{0.812}}}&{\textbf{\textcolor{black}{0.818}}}& {\textbf{\textcolor{black}{0.473}}}\\
\hline DeepWalk & 0.14 &0.6&0.56& 0.12\\
\hline node2vec & 0.14 & 0.56&0.52& 0.09\\
\hline RUM & {\textbf{\textcolor{blue}{0.16}}} & {\textbf{\textcolor{blue}{0.7}}}&{\textbf{\textcolor{blue}{0.58}}}& {\textbf{\textcolor{blue}{0.133}}}\\
\hline
\end{tabular}
\label{tbl:rc}
\end{table}
		
Since LINE1's mechanism focuses primarily and solely on preserving the first order proximity, a.k.a. direct links,  the embedding's goal is to approximate the original affinity matrix. This gives LINE1 substantial advantage for network reconstruction where the objective is to reconstruct the affinity matrix. We mark its results as bold text. Therefore we mainly compare the results with other random walk-based methods, and for each dataset mark the best result in blue. It is evident that 
RUM consistently outperforms DeepWalk and node2vec by a considerable margin: for email-EU-core,  CoRA, CiteSeer and BlogCatalog, RUM's MAP takes lead by $0.02$, $0.1$, $0.02$ and $0.013$, marking improvements by $14.3\%$, $16.7\%$, $3.6\%$, and $10.8\%$ perspectively. This shows that as a framework that aims to preserve the full structural information, RUM indeed captures the local relationships effectively, and also benefits from its neighborhood/community information in the reconstruction process.

					\begin{figure*}[htp]
			%%\vspace{-0.3cm}
			%\hspace{-0.5cm}
			\centering
			\subfigure[t-SNE on Affinity Matrix]{
				\hspace{-1.4cm}
				\includegraphics[scale = 0.35]{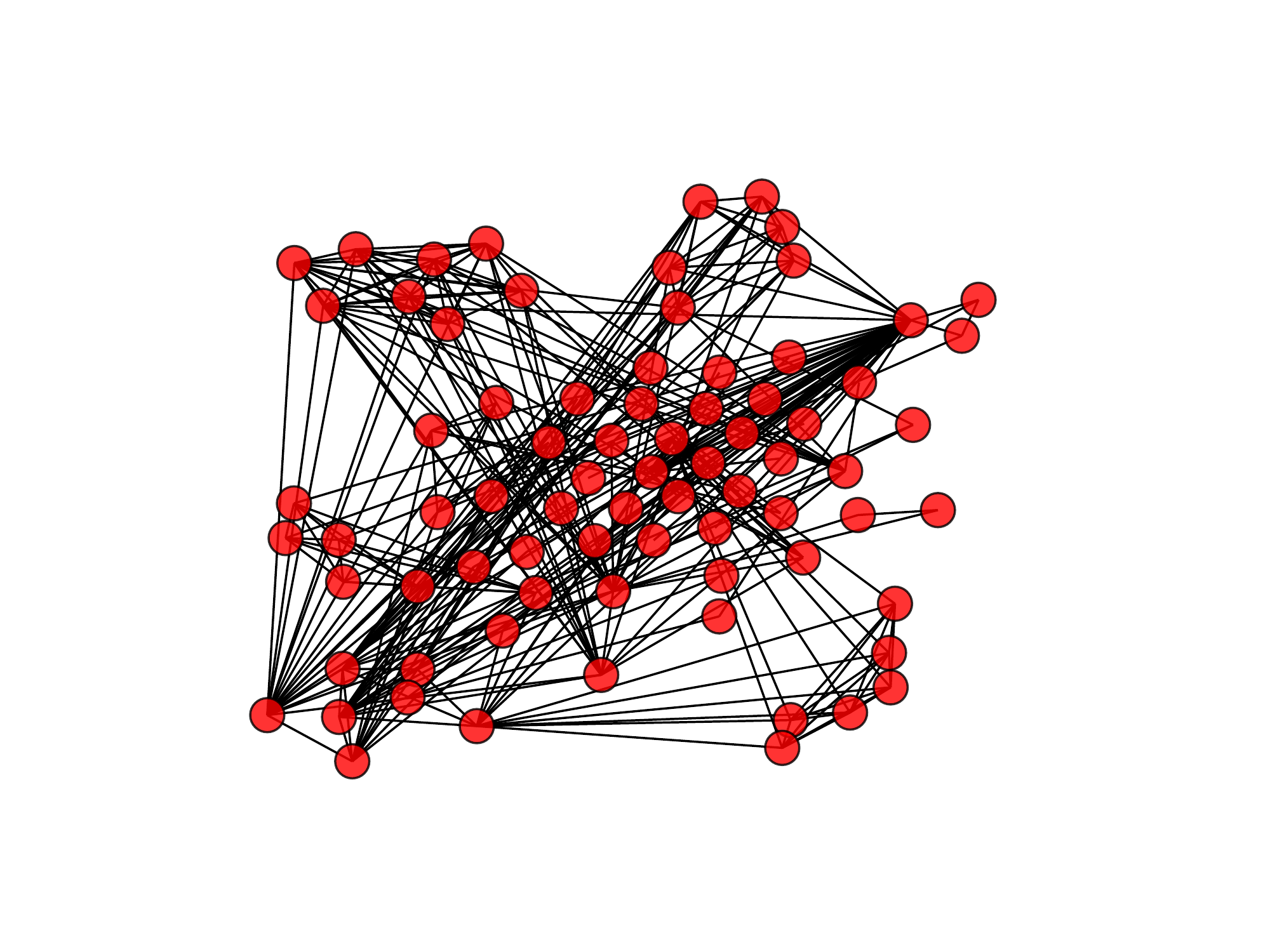}        \label{fig:vs1}
			}
			\subfigure[t-SNE on node2vec Embeddings]{
			 \hspace{-1.6cm}
				\includegraphics[scale = 0.35]{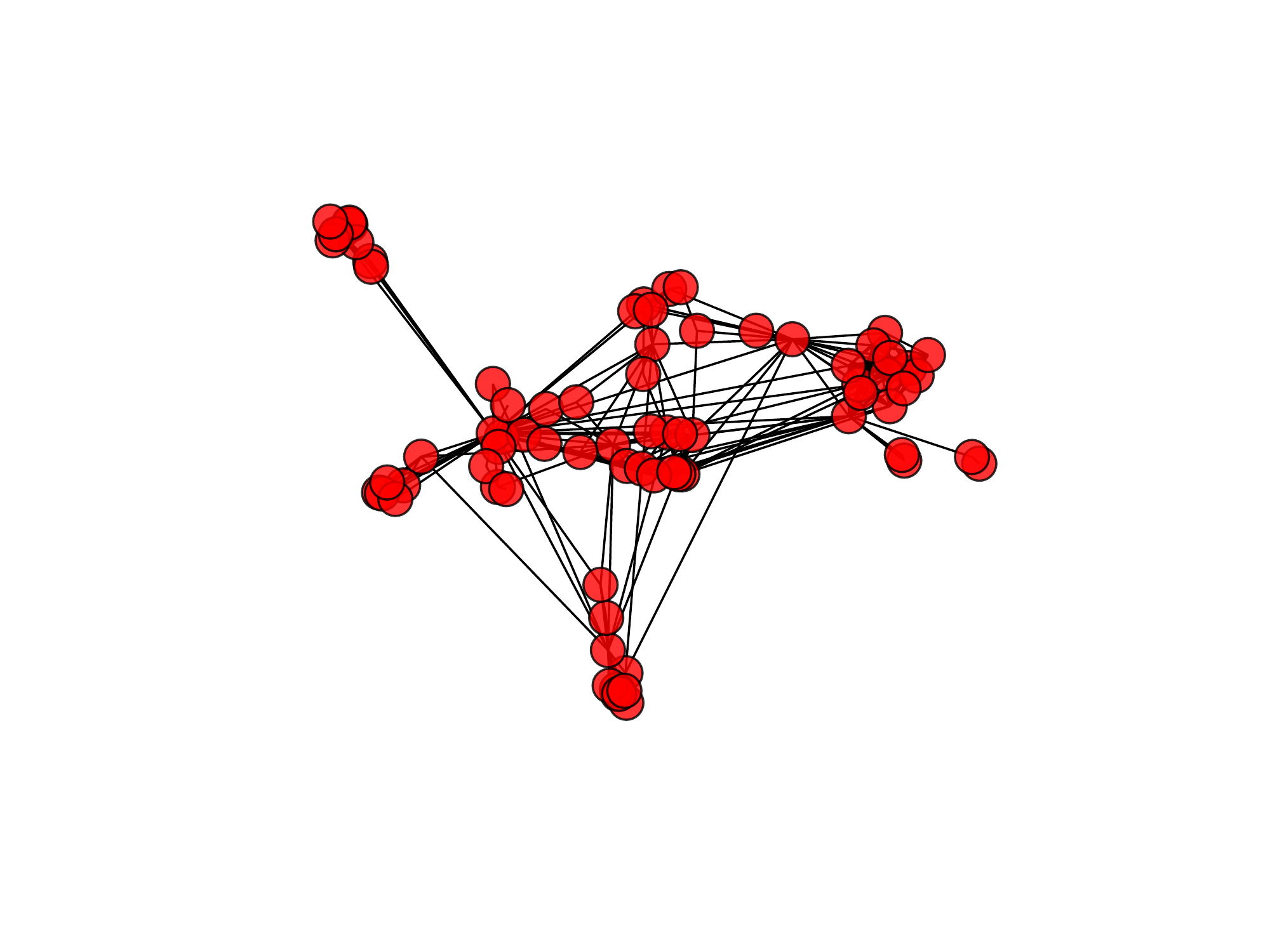}        \label{fig:vs2}
			}
			\subfigure[t-SNE on RUM Embeddings]{
			\hspace{-1.6cm}
				\includegraphics[scale = 0.35]{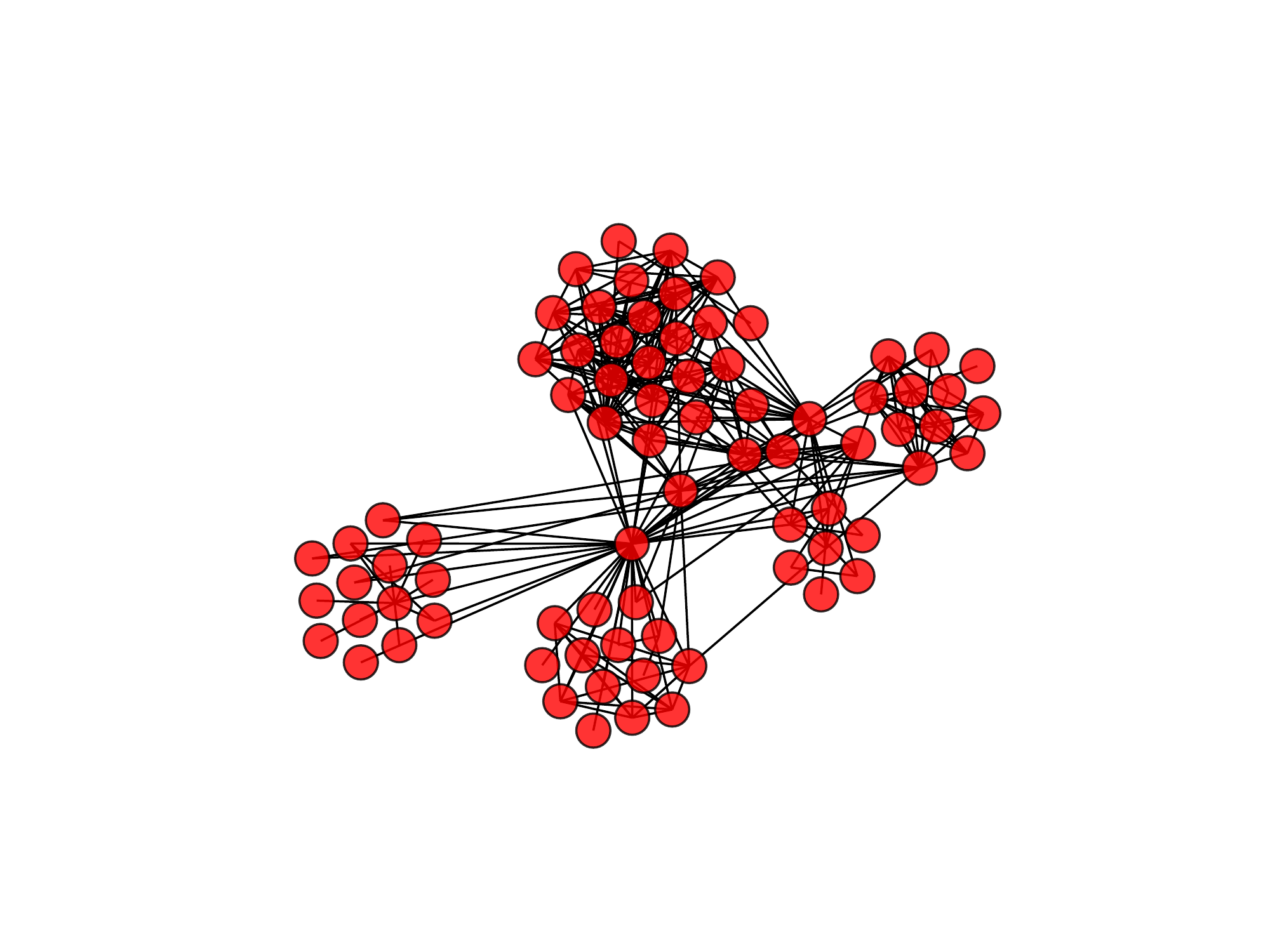}        \label{fig:vs3}
			}\\
			\subfigure[Affinity Propagation on Affinity Matrix]{
				\hspace{-1.4cm}
				\includegraphics[scale = 0.35]{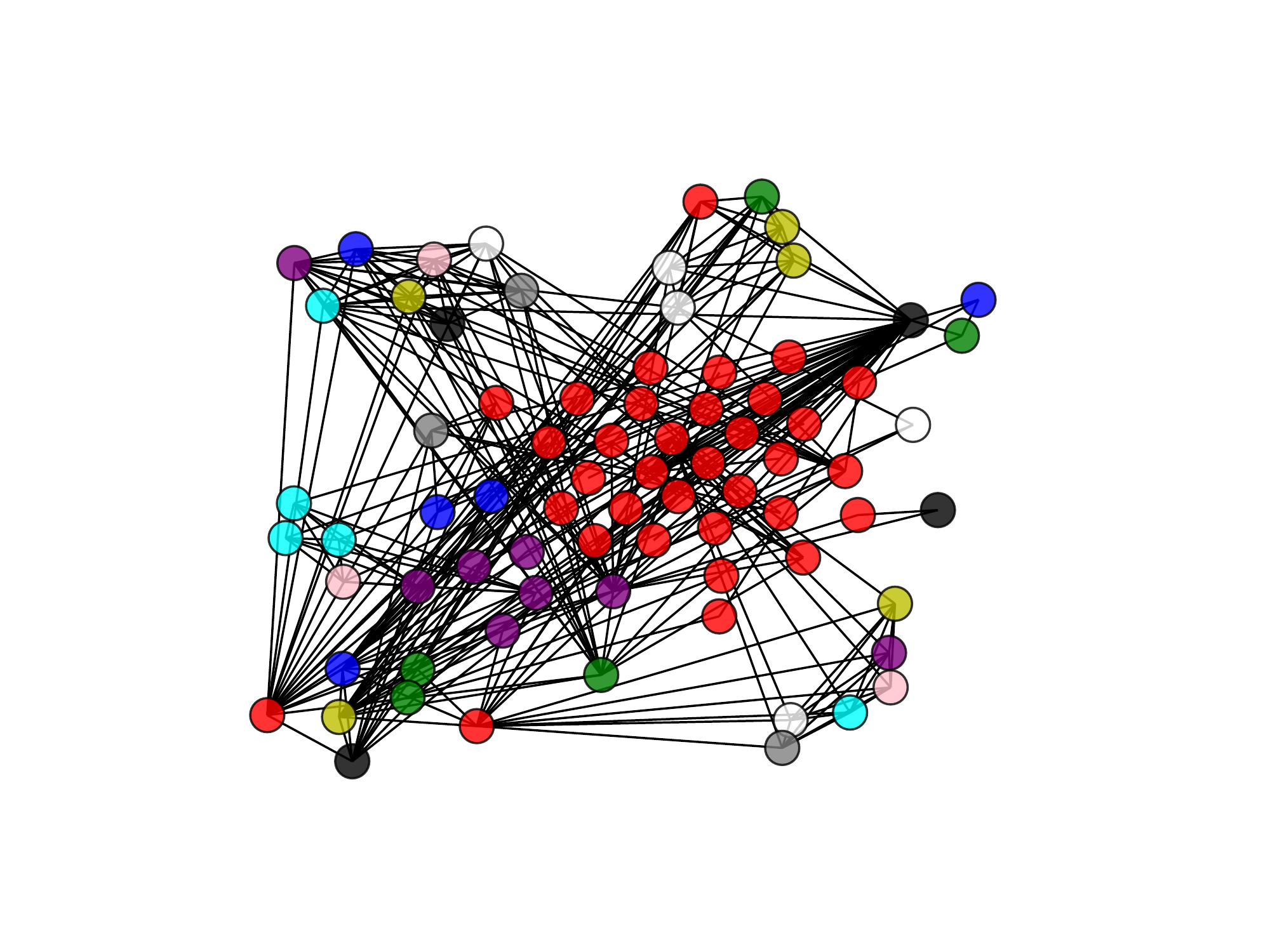}        \label{fig:vs4}
			}
			\subfigure[Affinity Propagation on node2vec Embeddings]{
			 \hspace{-1.6cm}
				\includegraphics[scale = 0.35]{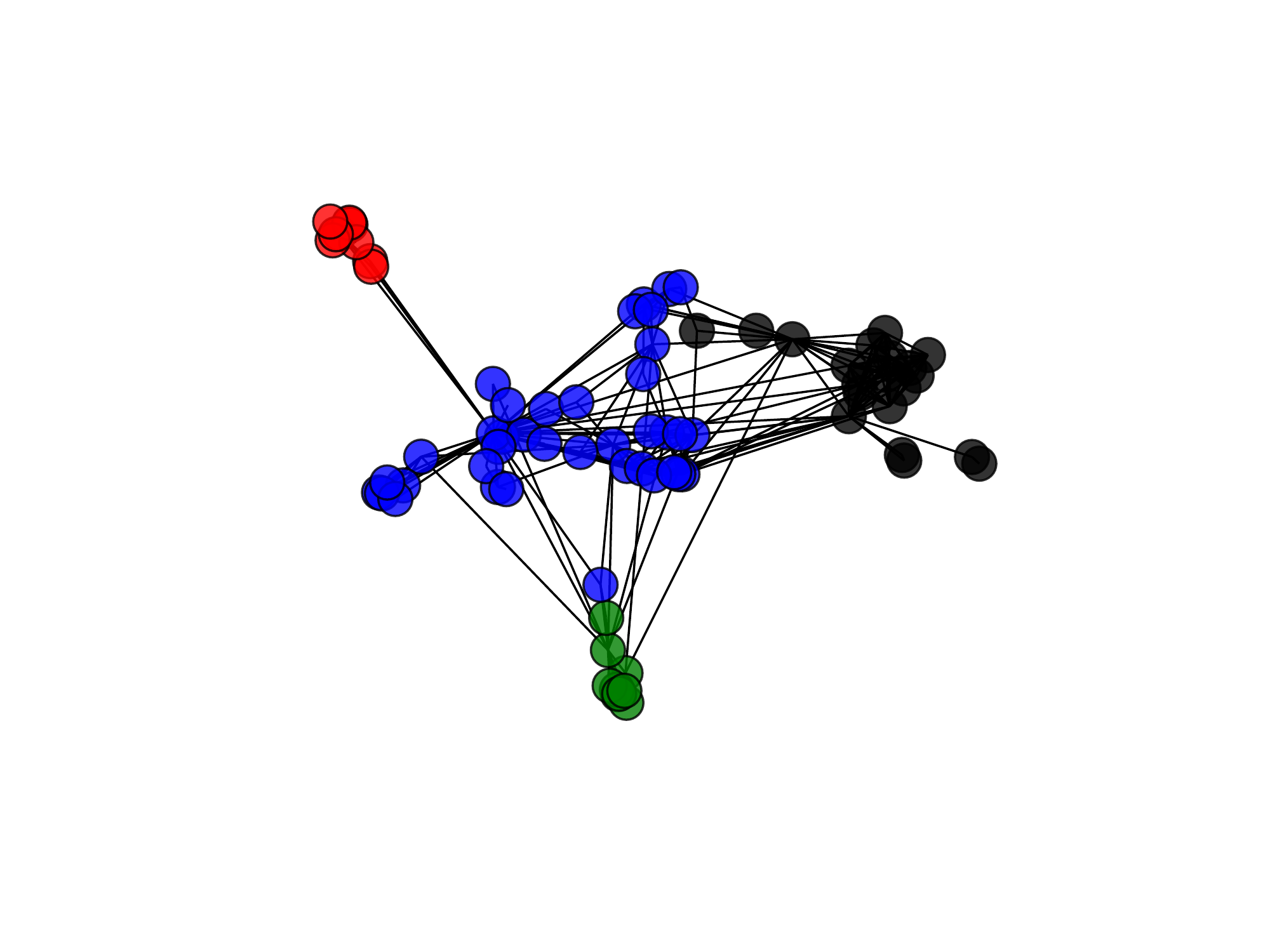}        \label{fig:vs5}
			}
			\subfigure[Affinity Propagation on RUM Embeddings]{
			\hspace{-1.6cm}
				\includegraphics[scale = 0.35]{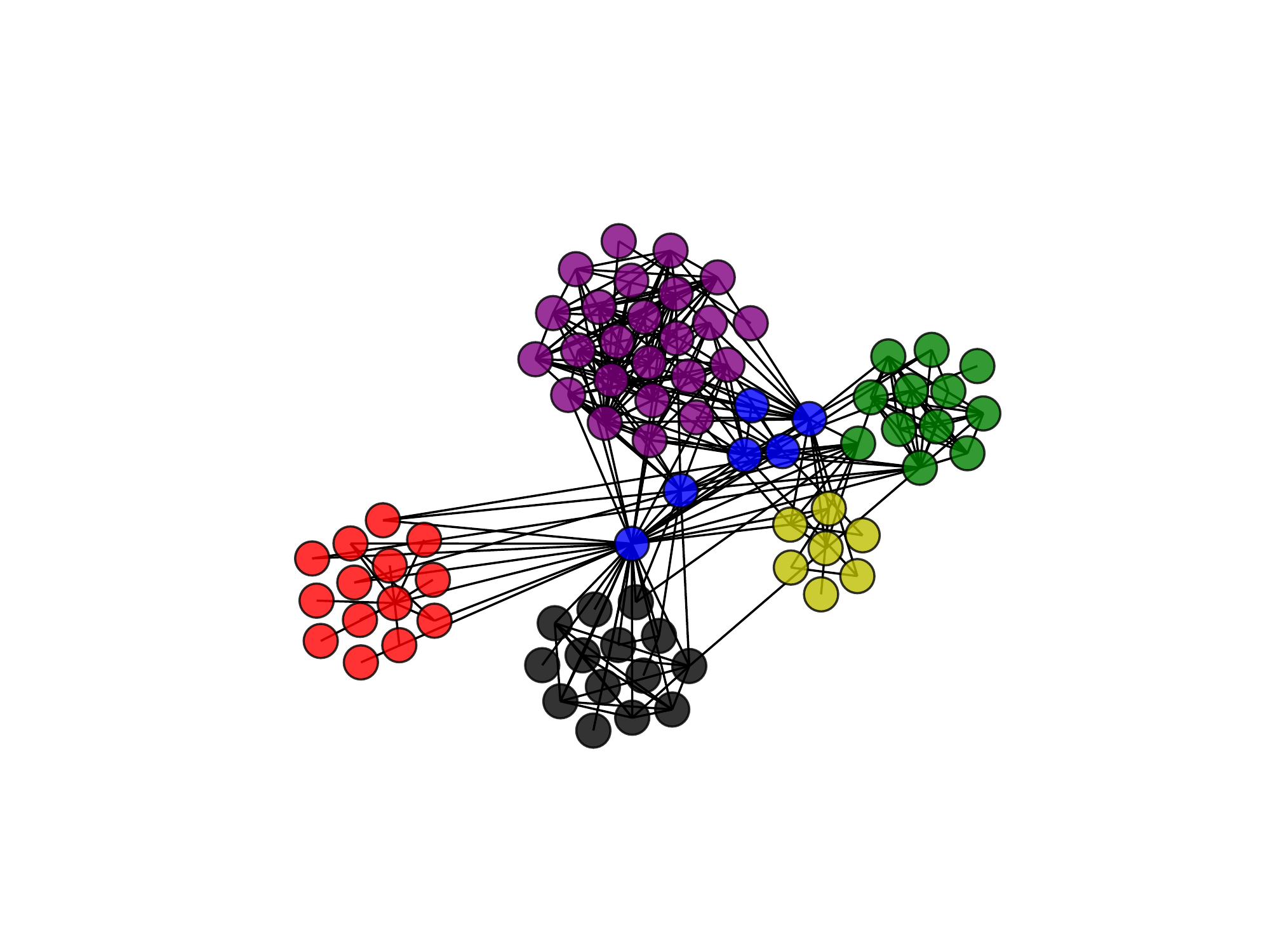}        \label{fig:vs6}
			}

			%%\vspace{-0.2cm}
			\caption{Visualization of the Les Mis\'erables network. } 
			\label{fig:vs}
			%%%\vspace{-0.6cm}
		\end{figure*}
	
	\subsection{Case Study}
		We present an interesting case study. Here we use the Les Mis\'erables network \footnote{\url{http://moreno.ss.uci.edu/data.html#lesmis}}, a network that is formed by connecting the novel characters by co-appearances. The story consists of 77 characters, and unique 254 pairs of them appeared in the same context, forming a network of 77 nodes and 254 edges. To show how differently our model work with existing methods, we first run our embedding model and generate a $16$-dimensional feature vector for each of the nodes, and then run t-SNE \cite{maaten2008visualizing} to further reduce its dimensionality to two, and use the $2$-d vectors as coordinates to depict the network structure. For comparison, we also demonstrate the visualization on the original co-appearance network and on the representations learned by node2vec \cite{grover2016node2vec} ($16$-d embedding vectors), which is a widely known embedding method developed in recent year. A comparative visualization of is shown in Figure \ref{fig:vs}. Figures \ref{fig:vs1}, \ref{fig:vs2}, \ref{fig:vs3} show the t-SNE visualization based on $2$-d t-SNE embeddings, generated from the original affinity matrix, the node2vec embedding vectors, the RUM embedding vectors perspectively. Figures \ref{fig:vs4}, \ref{fig:vs5}, \ref{fig:vs6} illustrate the results of Affinity Propagation clustering on affinity matrix, node2vec and RUM perspectively. The color on the nodes is determined by its cluster label.

		Here we have two observations. Firstly, when transformed to a low dimensional space, RUM embeddings are more easily separable, indicating that the structural information has been captured successfully by the RUM network representations. Secondly, zooming in on the clusters, we can see that the red, purple, yellow and green clusters demonstrate high level of homophily, while the blue cluster shows strong property of structural equivalence, i.e. though the blue nodes are not as tightly connected internally as its purple counterpart, it members do play the same role as a ``hub'' for other clusters. For example, both the two blue nodes at bottom left are connected directly to the same nodes in the purple and black clusters; while blue nodes at upper right receive connections from the purple and green clusters. It is evident that RUM generates very high quality representations.

	%%\vspace{-0.2cm}
	
	\section{Related Work}
	\label{sec:rw}
	Proposed in the early 2000s, Network Representation Learning (NRL) has attracted much research effort in recent years. Some earlier work \cite{Roweis:2000aa,Tenenbaum:2000aa,Belkin:2002aa} such as Local Linear Embedding (LLE)\cite{Roweis:2000aa}, IsoMA \cite{Tenenbaum:2000aa} and Laplacian Eigenma \cite{Belkin:2002aa} pioneered on the learning of low-dimension representation for graph nodes rather than sparse adjacent vector which requires more storage space and computing resources. These approaches make use of the linear transformations and embed the graph by solving the eigen problem of the adjacency matrix, Laplacian matrix, node transition probability matrix or other affinity matrix which represent the connection between nodes. The idea behind these methods is that if two nodes are connected, their embedding vectors should also be similar in the latent embedding space.
	
	The idea of network embedding gained attention in many research areas. In nature language processing, Word2vec \cite{NIPS2013_5021}, the widely recognized word embedding method, is developed in 2013 by Google.
	Word2vec learns vectors to keep words that have same context in close distances in the embedding space. It employs the skip-gram model to improve the training performance. Inspired by Word2vec, Perozzi et al proposed DeepWalk \cite{Perozzi:2014:DOL:2623330.2623732} for graph embedding, by uniform sampling rand walks in graphs to transform the graph structure into linear sequences of nodes, which are subsequently treated as sentences in language models and fed to the Skip-gram model to learn the graph embedding. 
	As DeepWalk generates walks randomly, node2vec \cite{grover2016node2vec} modifies the way of generating sequences and introduces two parameters to control random walk strategies between Breedth First Sampling (BFS) sampling and Depth First Sampling (DFS) . LINE \cite{Tang:2015:LLI:2736277.2741093} is the first algorithm to define an explicit objective function to preserve both first-order and second-order proximities between vertices. By introducing specific proximities that capture these graph properties, LINE achieved better performance. Inspired by the idea of exploiting multiple orders of proximity, GraRep \cite{DBLP:conf/cikm/CaoLX15} defines higher-order proximities and factorizes different k-order proximity matrices to obtain embedding vectors. Though GraRep suffers heavily from high computational costs, it achieves better performance on predictive tasks such as node classification. Its performance in empirical evaluation proves that preserving higher proximity indeed contributes better network embedding.In network embedding, random walk is widely accepted and \cite{DBLP:journals/corr/PerozziKS16} applies random walks to embed  graph nodes for network classification. \cite{DBLP:journals/corr/GoyalF17} summarized some exiting graph techniques,applications and performance, laiding the foundation for the follow-up study.
	
	In 2014, Levy and Goldberg proved that DeepWalk implicitly factorizes a matrix based on the node-node co-occurrence relationships \cite{levy2014neural}. Many have pointed out that some methods \cite{Tang:2015:LLI:2736277.2741093,Perozzi:2014:DOL:2623330.2623732}  only concentrate on first order and second order proximity, which is only based on local information. Others \cite{DBLP:journals/corr/ShahRD17,DBLP:conf/cikm/CaoLX15} have noticed that higher proximity should be considered to improve embedding quality. Yang, et al. \cite{DBLP:conf/ijcai/YangSLT17} designed a unified framework to cover these methods and promote the Network Embedding Updata (NEU) algorithm. NEU found a fast way to achieve high-proximity information on the simulation and improved existing network embedding algorithms. Similar to NEU, Chen, et al. \cite{DBLP:journals/corr/ChenNAKF17} have introduced a fast, warped graph embedding method and also designed a unifying framework to summarize the state-of-art methods. Meanwhile, some researchers \cite{tian2014learning,yang2016modularity,DBLP:conf/kdd/WangC016,DBLP:conf/ijcai/PanWZZW16,DBLP:journals/corr/ZhangWYWZ17,DBLP:conf/aaai/CaoLX16} promoted deep learning-based methods to improve network embedding. Meanwhile, there is research effort \cite{DBLP:conf/kdd/RibeiroSF17,DBLP:conf/aaai/LinLSLZ15} focusing on network property and trying to understand the network more thoroughly. For example, \cite{DBLP:conf/icdm/ChangQAZWH14} specifically analyzes the similarity of nodes in networks.
	
	Studies on high-order proximities have also encouraged the use of macroscopic information such as community in networks. \cite{DBLP:conf/wsdm/YangL13} finds an overlapping community detection method which is scalable to large networks. In 2016, zheng, et al. \cite{DBLP:journals/corr/ZhengCCCC16} proposed an algorithm from node embedding to community embedding, by introducing the concept of community in addition to nodes. The method jointly optimizes the node embedding and the community embedding, so that node embeddings also preserve the community structures. Similarly,  \cite{DBLP:journals/corr/TuWZLS16} also introduced the community structures to network embedding and showed that community structures capture vital global structural patterns. They proposed a method called Community-enhanced Network Representation Learning (CNRL) and obtained more informative representation with community information. Another network embedding method that considers the community structure is the Community Preserving Network Embedding \cite{DBLP:conf/aaai/WangCWP0Y17} method, which utilizes the M-NMF model to incorporate the community structure into network embeddings. By jointly considering  community structures and the first/second-order proximity, CPNE achieves competitive performance on network embedding and community detection. Our method also embeds network nodes with the  assistance of community structures. A major difference between our approach and existing methods is that existing methods tend to embed network nodes and find community structures simultaneously, while our method focuses on improving NRL quality by incorporating community information and other information that reside in multiple levels in a network.
	
	Compared to the algorithms above, most of which only investigate network structure, there are some novel trend forming up \cite{DBLP:conf/pakdd/ChenMLZL17,DBLP:conf/ijcai/YangLZSC15,DBLP:journals/corr/LiaoHZC17,DBLP:conf/dasfaa/LiWYLYZZ17,DBLP:conf/ijcai/YangLZSC15,DBLP:conf/ijcai/TuZLS17,DBLP:conf/acl/TuLLS17}, by considering other information in network to obtain better embeddings. \cite{DBLP:conf/kdd/OuCPZ016} showed that addition information besides network structures is also important. TAD \cite{DBLP:conf/ijcai/YangLZSC15} incorporates text features of vertices into network representation learning. \cite{DBLP:conf/acl/TuLLS17} proposed Context Aware Network Embedding(CANE) to model the semantic relationships more precisely by treating different neighbors with different reasons. TransNet \cite{DBLP:conf/ijcai/TuZLS17} utilizes rich semantic information on edges and regard the interactions between nodes as translation operation to conduct network embedding. Meanwhile, some research \cite{DBLP:conf/dasfaa/LiLW0ZZ17,DBLP:journals/corr/LiangJP17} apply the idea of semi-supervised to utilize inferred label information.
	
	Except a few embedding methods that focus on homogeneous network, there are research effort \cite{DBLP:conf/wsdm/JacobDG14,DBLP:conf/kdd/TangQM15,DBLP:conf/kdd/DongCS17,DBLP:conf/dasfaa/ChenW17} on heterogeneous network embedding. Predictive Text Embedding (PTE \cite{DBLP:conf/kdd/TangQM15} is a classic method which employs the semi-supervised idea to improve the network embedding and applies the embedding method to heterogeneous network. Metapath2vec \cite{DBLP:conf/kdd/DongCS17} also develops scalable representation learning for heterogeneous network, formalizing meta-path-based random walks and leveraging a heterogeneous skip-gram model. Yuxin Chen proposed HIN \cite{DBLP:conf/dasfaa/ChenW17} that carries out network embedding with both local and global information. 
	
	\section{Conclusion}
	\label{sec:conc}
	We propose a power and flexible model framework called RUM, for network representation by learning multiple layers of structural information in a network. RUM jointly considers three proximities, namely the triadic proximity, the global community proximity, and the neighborhood proximity. In other words, RUM is able to capture the finest details in the local relationships, while also keeping tracking of the larger groups of the global scale. RUM's learning strategy therefore covers the entire structural information spectrum of a network. The framework is also flexible and can take advantage of any newly developed community discovery algorithms and use them to improve the quality of network representations. In empirical evaluation, RUM has demonstrated advantages over state-of-the-art methods. In future we aim to improve the scalability of the framework, by parallelization and other optimizations to the current framework.
	
	%%\vspace{-0.35cm}

	\section*{Acknowledgment}
	\eat{This work was partially supported by the Fundamental Research Funds for the Central Universities, and the Research Funds of Renmin University of China (No.2015030275). The majority of the work was done while the first author was with the CSIRO, and with the support from the CSIRO Batmon project. }

	\bibliographystyle{abbrv}
	\bibliography{ICDE18_research_024}

\begin{thebibliography}{10}

\bibitem{Belkin:2002aa}
M.~Belkin, P.~Niyogi, T.~G. Dietterich, S.~Becker, and Z.~Ghahramani.
\newblock Laplacian eigenmaps and spectral techniques for embedding and
  clustering.
\newblock In {\em Advances In Neural Information Processing Systems 14, Vols 1
  and 2}, volume~14, pages 585--591, FIVE CAMBRIDGE CENTER, CAMBRIDGE, MA 02142
  USA, 2002. M I T PRESS.

\bibitem{DBLP:conf/cikm/CaoLX15}
S.~Cao, W.~Lu, and Q.~Xu.
\newblock Grarep: Learning graph representations with global structural
  information.
\newblock In {\em Proceedings of the 24th {ACM} International Conference on
  Information and Knowledge Management, {CIKM} 2015, Melbourne, VIC, Australia,
  October 19 - 23, 2015}, pages 891--900, 2015.

\bibitem{DBLP:conf/aaai/CaoLX16}
S.~Cao, W.~Lu, and Q.~Xu.
\newblock Deep neural networks for learning graph representations.
\newblock In {\em Proceedings of the Thirtieth {AAAI} Conference on Artificial
  Intelligence, February 12-17, 2016, Phoenix, Arizona, {USA.}}, pages
  1145--1152, 2016.

\bibitem{DBLP:conf/icdm/ChangQAZWH14}
S.~Chang, G.~Qi, C.~C. Aggarwal, J.~Zhou, M.~Wang, and T.~S. Huang.
\newblock Factorized similarity learning in networks.
\newblock In {\em 2014 {IEEE} International Conference on Data Mining, {ICDM}
  2014, Shenzhen, China, December 14-17, 2014}, pages 60--69, 2014.

\bibitem{DBLP:journals/corr/ChenNAKF17}
S.~Chen, S.~Niu, L.~Akoglu, J.~Kovacevic, and C.~Faloutsos.
\newblock Fast, warped graph embedding: Unifying framework and one-click
  algorithm.
\newblock {\em CoRR}, abs/1702.05764, 2017.

\bibitem{DBLP:conf/pakdd/ChenMLZL17}
W.~Chen, X.~Mao, X.~Li, Y.~Zhang, and X.~Li.
\newblock {PNE:} label embedding enhanced network embedding.
\newblock In {\em Advances in Knowledge Discovery and Data Mining - 21st
  Pacific-Asia Conference, {PAKDD} 2017, Jeju, South Korea, May 23-26, 2017,
  Proceedings, Part {I}}, pages 547--560, 2017.

\bibitem{DBLP:conf/dasfaa/ChenW17}
Y.~Chen and C.~Wang.
\newblock {HINE:} heterogeneous information network embedding.
\newblock In {\em Database Systems for Advanced Applications - 22nd
  International Conference, {DASFAA} 2017, Suzhou, China, March 27-30, 2017,
  Proceedings, Part {I}}, pages 180--195, 2017.

\bibitem{DBLP:conf/kdd/DongCS17}
Y.~Dong, N.~V. Chawla, and A.~Swami.
\newblock metapath2vec: Scalable representation learning for heterogeneous
  networks.
\newblock In {\em Proceedings of the 23rd {ACM} {SIGKDD} International
  Conference on Knowledge Discovery and Data Mining, Halifax, NS, Canada,
  August 13 - 17, 2017}, pages 135--144, 2017.

\bibitem{DBLP:journals/corr/GoyalF17}
P.~Goyal and E.~Ferrara.
\newblock Graph embedding techniques, applications, and performance: {A}
  survey.
\newblock {\em CoRR}, abs/1705.02801, 2017.

\bibitem{grover2016node2vec}
A.~Grover and J.~Leskovec.
\newblock node2vec: Scalable feature learning for networks.
\newblock In {\em Proceedings of the 22nd ACM SIGKDD international conference
  on Knowledge discovery and data mining}, pages 855--864. ACM, 2016.

\bibitem{DBLP:conf/wsdm/JacobDG14}
Y.~Jacob, L.~Denoyer, and P.~Gallinari.
\newblock Learning latent representations of nodes for classifying in
  heterogeneous social networks.
\newblock In {\em Seventh {ACM} International Conference on Web Search and Data
  Mining, {WSDM} 2014, New York, NY, USA, February 24-28, 2014}, pages
  373--382, 2014.

\bibitem{Leck:2000aa}
R.~M. Leck.
\newblock {\em Georg Simmel and avant-garde sociology: the birth of modernity,
  1880-1920}.
\newblock Humanity Books, Amherst, N.Y., 2000.

\bibitem{levy2014neural}
O.~Levy and Y.~Goldberg.
\newblock Neural word embedding as implicit matrix factorization.
\newblock In {\em Advances in neural information processing systems}, pages
  2177--2185, 2014.

\bibitem{DBLP:conf/dasfaa/LiLW0ZZ17}
C.~Li, Z.~Li, S.~Wang, Y.~Yang, X.~Zhang, and J.~Zhou.
\newblock Semi-supervised network embedding.
\newblock In {\em Database Systems for Advanced Applications - 22nd
  International Conference, {DASFAA} 2017, Suzhou, China, March 27-30, 2017,
  Proceedings, Part {I}}, pages 131--147, 2017.

\bibitem{DBLP:conf/dasfaa/LiWYLYZZ17}
C.~Li, S.~Wang, D.~Yang, Z.~Li, Y.~Yang, X.~Zhang, and J.~Zhou.
\newblock {PPNE:} property preserving network embedding.
\newblock In {\em Database Systems for Advanced Applications - 22nd
  International Conference, {DASFAA} 2017, Suzhou, China, March 27-30, 2017,
  Proceedings, Part {I}}, pages 163--179, 2017.

\bibitem{DBLP:journals/corr/LiangJP17}
J.~Liang, P.~Jacobs, and S.~Parthasarathy.
\newblock {SEANO:} semi-supervised embedding in attributed networks with
  outliers.
\newblock {\em CoRR}, abs/1703.08100, 2017.

\bibitem{DBLP:journals/corr/LiaoHZC17}
L.~Liao, X.~He, H.~Zhang, and T.~Chua.
\newblock Attributed social network embedding.
\newblock {\em CoRR}, abs/1705.04969, 2017.

\bibitem{DBLP:conf/aaai/LinLSLZ15}
Y.~Lin, Z.~Liu, M.~Sun, Y.~Liu, and X.~Zhu.
\newblock Learning entity and relation embeddings for knowledge graph
  completion.
\newblock In {\em Proceedings of the Twenty-Ninth {AAAI} Conference on
  Artificial Intelligence, January 25-30, 2015, Austin, Texas, {USA.}}, pages
  2181--2187, 2015.

\bibitem{maaten2008visualizing}
L.~v.~d. Maaten and G.~Hinton.
\newblock Visualizing data using t-sne.
\newblock {\em Journal of Machine Learning Research}, 9(Nov):2579--2605, 2008.

\bibitem{NIPS2013_5021}
T.~Mikolov, I.~Sutskever, K.~Chen, G.~S. Corrado, and J.~Dean.
\newblock Distributed representations of words and phrases and their
  compositionality.
\newblock In C.~J.~C. Burges, L.~Bottou, M.~Welling, Z.~Ghahramani, and K.~Q.
  Weinberger, editors, {\em Advances in Neural Information Processing Systems
  26}, pages 3111--3119. Curran Associates, Inc., 2013.

\bibitem{DBLP:conf/kdd/OuCPZ016}
M.~Ou, P.~Cui, J.~Pei, Z.~Zhang, and W.~Zhu.
\newblock Asymmetric transitivity preserving graph embedding.
\newblock In {\em Proceedings of the 22nd {ACM} {SIGKDD} International
  Conference on Knowledge Discovery and Data Mining, San Francisco, CA, USA,
  August 13-17, 2016}, pages 1105--1114, 2016.

\bibitem{DBLP:conf/ijcai/PanWZZW16}
S.~Pan, J.~Wu, X.~Zhu, C.~Zhang, and Y.~Wang.
\newblock Tri-party deep network representation.
\newblock In {\em Proceedings of the Twenty-Fifth International Joint
  Conference on Artificial Intelligence, {IJCAI} 2016, New York, NY, USA, 9-15
  July 2016}, pages 1895--1901, 2016.

\bibitem{Perozzi:2014:DOL:2623330.2623732}
B.~Perozzi, R.~Al-Rfou, and S.~Skiena.
\newblock Deepwalk: Online learning of social representations.
\newblock In {\em Proceedings of the 20th ACM SIGKDD International Conference
  on Knowledge Discovery and Data Mining}, KDD '14, pages 701--710, New York,
  NY, USA, 2014. ACM.

\bibitem{DBLP:journals/corr/PerozziKS16}
B.~Perozzi, V.~Kulkarni, and S.~Skiena.
\newblock Walklets: Multiscale graph embeddings for interpretable network
  classification.
\newblock {\em CoRR}, abs/1605.02115, 2016.

\bibitem{DBLP:conf/kdd/RibeiroSF17}
L.~F.~R. Ribeiro, P.~H.~P. Saverese, and D.~R. Figueiredo.
\newblock \emph{struc2vec}: Learning node representations from structural
  identity.
\newblock In {\em Proceedings of the 23rd {ACM} {SIGKDD} International
  Conference on Knowledge Discovery and Data Mining, Halifax, NS, Canada,
  August 13 - 17, 2017}, pages 385--394, 2017.

\bibitem{Roweis:2000aa}
S.~T. Roweis and L.~K. Saul.
\newblock Nonlinear dimensionality reduction by locally linear embedding.
\newblock {\em Science}, 290(5500):2323--+, December 2000.

\bibitem{DBLP:journals/corr/ShahRD17}
V.~Shah, N.~Rao, and W.~Ding.
\newblock Matrix factorization with side and higher order information.
\newblock {\em CoRR}, abs/1705.02047, 2017.

\bibitem{DBLP:conf/kdd/TangQM15}
J.~Tang, M.~Qu, and Q.~Mei.
\newblock {PTE:} predictive text embedding through large-scale heterogeneous
  text networks.
\newblock In {\em Proceedings of the 21th {ACM} {SIGKDD} International
  Conference on Knowledge Discovery and Data Mining, Sydney, NSW, Australia,
  August 10-13, 2015}, pages 1165--1174, 2015.

\bibitem{Tang:2015:LLI:2736277.2741093}
J.~Tang, M.~Qu, M.~Wang, M.~Zhang, J.~Yan, and Q.~Mei.
\newblock Line: Large-scale information network embedding.
\newblock In {\em Proceedings of the 24th International Conference on World
  Wide Web}, WWW '15, pages 1067--1077, Republic and Canton of Geneva,
  Switzerland, 2015. International World Wide Web Conferences Steering
  Committee.

\bibitem{Tenenbaum:2000aa}
J.~B. Tenenbaum, V.~de~Silva, and J.~C. Langford.
\newblock A global geometric framework for nonlinear dimensionality reduction.
\newblock {\em Science}, 290(5500):2319--+, December 2000.

\bibitem{tian2014learning}
F.~Tian, B.~Gao, Q.~Cui, E.~Chen, and T.-Y. Liu.
\newblock Learning deep representations for graph clustering.
\newblock In {\em AAAI}, pages 1293--1299, 2014.

\bibitem{DBLP:conf/acl/TuLLS17}
C.~Tu, H.~Liu, Z.~Liu, and M.~Sun.
\newblock {CANE:} context-aware network embedding for relation modeling.
\newblock In {\em Proceedings of the 55th Annual Meeting of the Association for
  Computational Linguistics, {ACL} 2017, Vancouver, Canada, July 30 - August 4,
  Volume 1: Long Papers}, pages 1722--1731, 2017.

\bibitem{DBLP:journals/corr/TuWZLS16}
C.~Tu, H.~Wang, X.~Zeng, Z.~Liu, and M.~Sun.
\newblock Community-enhanced network representation learning for network
  analysis.
\newblock {\em CoRR}, abs/1611.06645, 2016.

\bibitem{DBLP:conf/ijcai/TuZLS17}
C.~Tu, Z.~Zhang, Z.~Liu, and M.~Sun.
\newblock Transnet: Translation-based network representation learning for
  social relation extraction.
\newblock In {\em Proceedings of the Twenty-Sixth International Joint
  Conference on Artificial Intelligence, {IJCAI} 2017, Melbourne, Australia,
  August 19-25, 2017}, pages 2864--2870, 2017.

\bibitem{DBLP:conf/kdd/WangC016}
D.~Wang, P.~Cui, and W.~Zhu.
\newblock Structural deep network embedding.
\newblock In {\em Proceedings of the 22nd {ACM} {SIGKDD} International
  Conference on Knowledge Discovery and Data Mining, San Francisco, CA, USA,
  August 13-17, 2016}, pages 1225--1234, 2016.

\bibitem{DBLP:conf/aaai/WangCWP0Y17}
X.~Wang, P.~Cui, J.~Wang, J.~Pei, W.~Zhu, and S.~Yang.
\newblock Community preserving network embedding.
\newblock In {\em Proceedings of the Thirty-First {AAAI} Conference on
  Artificial Intelligence, February 4-9, 2017, San Francisco, California,
  {USA.}}, pages 203--209, 2017.

\bibitem{DBLP:conf/ijcai/YangLZSC15}
C.~Yang, Z.~Liu, D.~Zhao, M.~Sun, and E.~Y. Chang.
\newblock Network representation learning with rich text information.
\newblock In {\em Proceedings of the Twenty-Fourth International Joint
  Conference on Artificial Intelligence, {IJCAI} 2015, Buenos Aires, Argentina,
  July 25-31, 2015}, pages 2111--2117, 2015.

\bibitem{DBLP:conf/ijcai/YangSLT17}
C.~Yang, M.~Sun, Z.~Liu, and C.~Tu.
\newblock Fast network embedding enhancement via high order proximity
  approximation.
\newblock In {\em Proceedings of the Twenty-Sixth International Joint
  Conference on Artificial Intelligence, {IJCAI} 2017, Melbourne, Australia,
  August 19-25, 2017}, pages 3894--3900, 2017.

\bibitem{DBLP:conf/wsdm/YangL13}
J.~Yang and J.~Leskovec.
\newblock Overlapping community detection at scale: a nonnegative matrix
  factorization approach.
\newblock In {\em Sixth {ACM} International Conference on Web Search and Data
  Mining, {WSDM} 2013, Rome, Italy, February 4-8, 2013}, pages 587--596, 2013.

\bibitem{yang2016modularity}
L.~Yang, X.~Cao, D.~He, C.~Wang, X.~Wang, and W.~Zhang.
\newblock Modularity based community detection with deep learning.
\newblock In {\em IJCAI}, pages 2252--2258, 2016.

\bibitem{Yin:2017:LHG:3097983.3098069}
H.~Yin, A.~R. Benson, J.~Leskovec, and D.~F. Gleich.
\newblock Local higher-order graph clustering.
\newblock In {\em Proceedings of the 23rd ACM SIGKDD International Conference
  on Knowledge Discovery and Data Mining}, KDD '17, pages 555--564, New York,
  NY, USA, 2017. ACM.

\bibitem{DBLP:journals/corr/ZhangWYWZ17}
W.~Zhang, L.~Wang, J.~Yan, X.~Wang, and H.~Zha.
\newblock Deep extreme multi-label learning.
\newblock {\em CoRR}, abs/1704.03718, 2017.

\bibitem{DBLP:journals/corr/ZhengCCCC16}
V.~W. Zheng, S.~Cavallari, H.~Cai, K.~C. Chang, and E.~Cambria.
\newblock From node embedding to community embedding.
\newblock {\em CoRR}, abs/1610.09950, 2016.

\end{thebibliography}
	
\end{document}